\documentclass[runningheads]{llncs}

\usepackage{times}
\usepackage{epsfig}
\usepackage{graphicx}
\usepackage{amsmath}
\usepackage{amssymb}
\usepackage{bm}
\usepackage{color, soul}
\usepackage[misc]{ifsym}

\definecolor{sarahcolor}{rgb}{0.858,0.188,0.478}

\begin{document}
\title{Probabilistic 3D surface reconstruction from sparse MRI information}
\titlerunning{Probabilistic 3D surface reconstruction from sparse MRI information}

%
\author{Katar\'{i}na T\'{o}thov\'{a}\inst{1}\orcidID{0000-0001-5864-179X} \** \and Sarah Parisot\inst{2} \and Matthew Lee\inst{3} \and Esther Puyol-Ant\'{o}n\inst{4}\orcidID{0000-0002-9965-7015} \and Andrew King\inst{4} \and Marc Pollefeys \inst{1,5} \and Ender Konukoglu\inst{1}}

%
%


\institute{ETH Zurich, Switzerland \\
\** \email{katarina.tothova@inf.ethz.ch}
\and Huawei Noah's Ark Lab, London, United Kingdom \\
\and Imperial College London, United Kingdom \\
\and King's College London, United Kingdom \\
\and Microsoft Mixed Reality and AI lab, Zurich, Switzerland}
\authorrunning{Katar\'{i}na T\'{o}thov\'{a} et al.}
\maketitle              
\begin{abstract}

Surface reconstruction from magnetic resonance (MR) imaging data is indispensable in medical image analysis and clinical research. A reliable and effective reconstruction tool should: be fast in prediction of accurate well localised and high resolution models, evaluate prediction uncertainty, work with as little input data as possible. Current deep learning state of the art (SOTA) 3D reconstruction methods, however, often only produce shapes of limited variability positioned in a canonical position or lack uncertainty evaluation. In this paper, we present a novel probabilistic deep learning approach for concurrent 3D surface reconstruction from sparse 2D MR image data and aleatoric uncertainty prediction. Our method is capable of reconstructing large surface meshes from three quasi-orthogonal MR imaging slices from limited training sets whilst modelling the location of each mesh vertex through a Gaussian distribution. Prior shape information is encoded using a built-in linear principal component analysis (PCA) model. Extensive experiments on cardiac MR data show that our probabilistic approach successfully assesses prediction uncertainty while at the same time qualitatively and quantitatively outperforms SOTA methods in shape prediction. Compared to SOTA, we are capable of properly localising and orientating the prediction via the use of a spatially aware neural network.

\end{abstract}

\section{Introduction}
High quality 3D surface models of internal organs constructed from MR imaging data are vital for diagnosis, disease tracking, surgical planning or interpretation of functional data in clinical and research practice \cite{peressutti2017,puyol2017,bai2018}. 
In cardiac imaging, for example, virtual ventricle surface meshes enable the use of patient-specific 3D models for investigation of valve and vessel function, or surgical and catheter-based procedural planning \cite{3Dprinting}.

The problem of surface reconstruction has been widely studied in medical imaging research. Traditional approaches usually take advantage of parametric models through atlas or statistical shape model registration \cite{puyol2017,peressutti2017,garcia-barnes2010,bai2013,denis2019} or use predefined forces to evolve a deformable shape into the final surface \cite{fischl2012,huo2016,han2004,schuh2017}. In contrast to the complex frameworks that might be associated with such methods such as in \cite{fischl2012}, the advent of machine learning has led to the possibility of training deep neural networks in an end-to-end manner: from images to parametrised shapes. In their work on 2D shape reconstruction \cite{milletari2017,workshop,bhalodia2018} employ convolutional neural networks (CNN) to modulate shapes generated from in-built principal component analysis (PCA) shape priors. The situation in \textit{3D} machine learning surface reconstruction from medical images is slightly more complex. Besides the usual requirements on accuracy and availability of sufficiently large training sets, the 3D methods often have to deal with sparsity of the input imaging information---as is the case in 3D reconstruction of organs from a few imaging slices. Support for sparse input data is essential in the medical domain. It leads to faster acquisition times, fewer motion artifacts, less radiation exposure for a patient (e.g. for CT), and ultimately a cheaper and more accessible imaging method. Cerrolaza et al. \cite{juan2018} solve the problem by relating it to the one of a single-view 3D reconstruction in general computer vision. When constructing volumetric predictions of 3D fetal skulls from 2D ultrasound images, they utilise a conditional variational autoencoder (CVAE) via a TL-net inspired architecture \cite{TL-net}. The downsides of such encoder-decoder setups are, however, computational and memory demands associated with computing 3D convolutions on volumetric data and limited resolution of predicted surfaces. Moreover, to build an implicit shape prior they require volumetric training data to be pre-aligned and hence at test time predict 3D shapes in a canonical position and orientation. The topology of predicted shapes is not constrained, which may lead to undesirable artefacts. Finally, Tatarchenko et al. \cite{tatarchenko} suggest there is in general little statistical difference between the reconstructions of encoder-decoder methods and shapes obtained as nearest-neighbours or local cluster means in the training set.

A prediction system that is to be useful in practice should be not only highly accurate and precise, it should also indicate how confident it is about the results it provides. This is of utmost interest in medical imaging due to the nature of the task at hand, its application and the input data, which is often times sparse and comprised of noisy images of coarse resolution riddled with imaging artifacts. The ambiguity ingrained in the data---\textit{aleatoric} uncertainty---is modelled via a probability distribution over the model output, which can be integrated into the optimisation process itself \cite{kendall2017,baumgartner2019}. In~\cite{workshop}, T\'{o}thov\'{a} et al. take this route by formulating 2D surface reconstruction as a conditional probability estimation problem. Another popular approach to assess the variability of the output of a method is the use of statistics aggregated on a set of plausible predictions sampled from a probabilistic model. Methods such as a conditional variational autoencoders \cite{juan2018}, MCMC \cite{chang2011,denis2019} or Gaussian processes \cite{le2016} have been used to this end.


In this paper, we propose a novel probabilistic model for 3D surface reconstruction from MRI data that jointly addresses high resolution accurate reconstruction, sparse input data constraints and aleatoric uncertainty estimation within a single framework. We learn to reconstruct high resolution surface meshes from three quasi-orthogonal MR imaging slices through the object using a small sized training dataset. Our method extends the PCA shape prior based surface prediction of \cite{milletari2017,workshop} into the probabilistic 3D setting by formulating the problem as a conditional probability estimation. Using a shape prior allows us to be robust in the absence of input information whilst our probabilistic formulation leads to an analytical expression capturing uncertainty. In contrast to autoencoder reconstruction methods \cite{juan2018,TL-net}, no training data pre-alignment is needed and model predictions are not only of the right shape, they are also correctly orientated, as shape orientation is predicted as a mode of the PCA model, and localised in the common world space. This is possible by augmenting the input features using coordinate maps relating the image and world space coordinate systems, a novel feature inspired by \cite{coordconv}. This approach also provides a simple yet efficient solution for dealing with possible heterogeneity in the input image acquisition setups, such as varying acquisition angles or positioning of patients in the scanner. The proposed approach was evaluated on 3D cardiac reconstruction using UK BioBank \cite{UKBB} MRI data. Our approach successfully assesses prediction uncertainty, outperforming  SOTA methods \cite{milletari2017,TL-net,juan2018} in terms of quantitative and qualitative evaluation. To our knowledge, we are the first to tackle the combined problem of reconstructing high resolution meshes in common world coordinates while preserving shape volume and topology and at the same time providing a principled quantification of uncertainty.

\section{Method}

Our goal is to devise a probabilistic 3D surface reconstruction model predicting organ surface meshes from a sparse set of 2D MRI input images. Specifically, we consider an input set of three MRI quasi-orthogonal slices across an organ denoted as $\bm{x}=\{\bm{X_1}, \bm{X_2}, \bm{X_3}\}$. Based on principles of probabilistic PCA \cite{bishop}, our framework addresses the inherent challenges linked with the sparse and heterogeneous input data via the use of a spatially aware deep CNN computing distributions over principal component scores of a PCA shape prior as in \cite{workshop}.
\begin{figure}[t]
    \centering
    \includegraphics[scale=0.35]{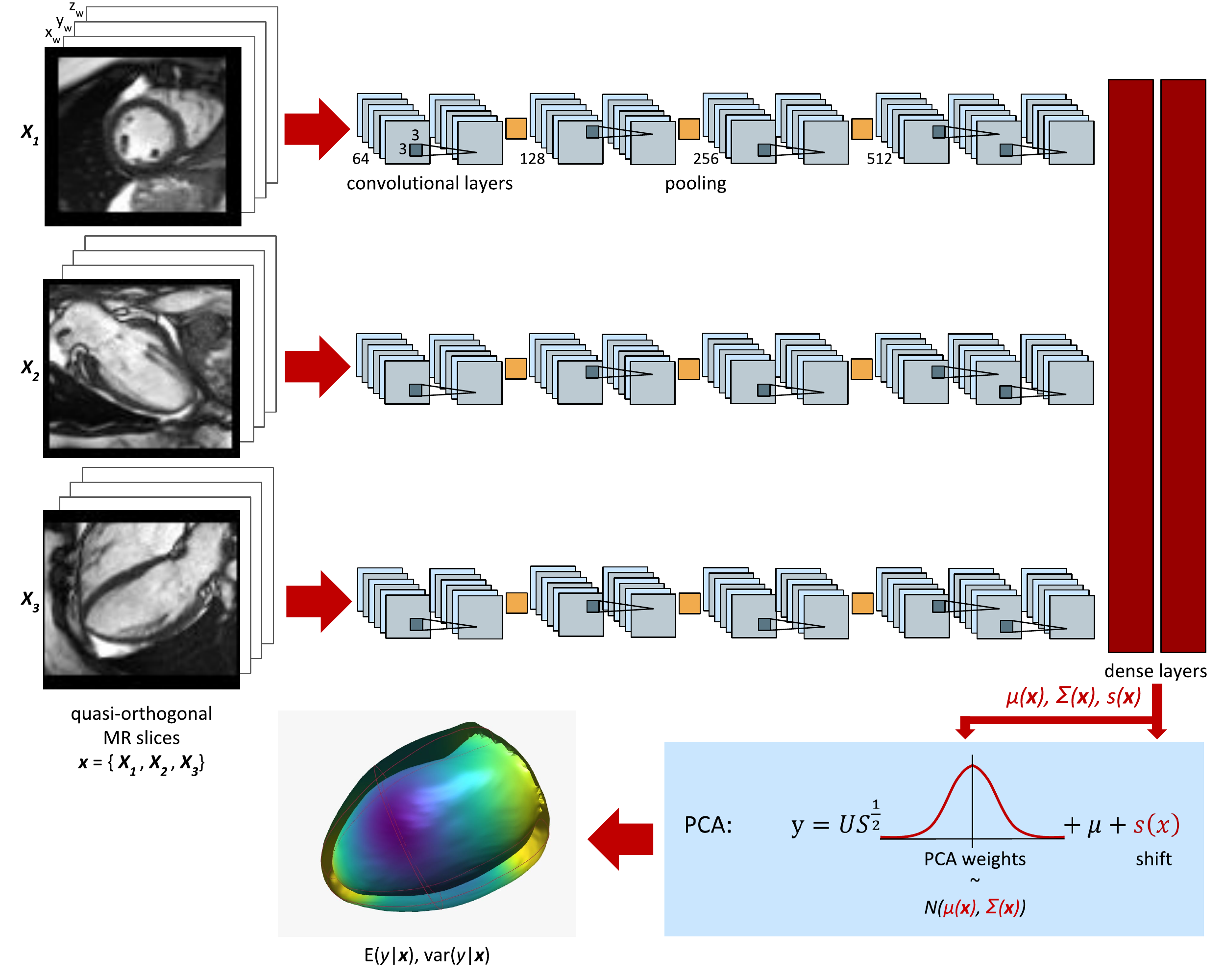}
    \caption{Schematic setup of the surface and uncertainty prediction. The input to each of the three convolutional branches of the network is a single MRI slice (here pictured cardiac MRI) concatenated with the coordinate maps computed from the metadata and relating the pixel and world space. The output of the framework is a probabilistic mesh where the location of each vertex is modelled by a Gaussian distribution.    }
    \label{fig:framework}
\end{figure}

\subsubsection{Probabilistic Model}
We formulate surface prediction as a probability estimation problem where we aim to compute a probability distribution $p(y|\bm{x})$ over the coordinates of surface mesh vertices $y$.%

Probabilities are conditioned on the input MRI imaging stack $\bm{x}$ through a latent variable model
\begin{equation}\label{pyx}
    p(y|\bm{x}) = \int p(y|z,\bm{x}) p(z|\bm{x}) dz,
\end{equation}
with latent variable $z$. For a given surface $y$, $z$ are scores in the PCA shape prior which we define through $p(y|z, \bm{x})$ as 
\begin{equation}\label{pyz}
    p(y|z,\bm{x}) = \mathcal{N} ( y | U S^{\frac{1}{2}}z + \mu + s(\bm{x}), {\sigma}^2 I),
\end{equation}
where $U$ is a matrix of principal vectors (columns of $U$),  $S$ represents the principal component diagonal covariance matrix and $\mu$ the data mean, all three precomputed using the surfaces in the training set, and $s$ is a global spatial shift dependent on the input image stack $\bm{x}$. Variance $\sigma^2$ reflects the noise level in the data. The latent space is conditioned on the input and structured according to the Gaussian distribution with
\begin{equation}\label{pzx}
	p(z|\bm{x}) = \mathcal{N}(z|\mu(\bm{x}), \Sigma(\bm{x})),
\end{equation}
where the mean $\mu(\bm{x})$ and covariance matrix $\Sigma(\bm{x})$ are estimated jointly from our input image stack $\bm{x}$ using a deep CNN, see Fig.~\ref{fig:framework}.  Note that in practice, we constrain the network to predict Cholesky factor $\Sigma^{\frac{1}{2}}(\bm{x})$ to ensure the positive definiteness of the estimated covariance matrix. Finally, a prior on the latent variable assumes $z\sim \mathcal{N}(0,I)$ such as in probabilistic PCA \cite{bishop}.

\subsubsection{Inference}
At test time, given a test image stack $\bm{x}$, we follow Eq. \ref{pyx} and generate the target mesh predictions $y$ by sampling from a Gaussian distribution $p(y|\bm{x}) = \mathcal{N}(\mathbb{E}(y|\bm{x}),\mathrm{var}(y|\bm{x}))$ with mean and variance expressed by
\begin{align}
	\mathbb{E}(y|\bm{x}) & = U S^{\frac{1}{2}} \mu(\bm{x}) + \mu + s, \label{eq:mean}\\ 
    \mathrm{var}(y|\bm{x}) & = \sigma^2I + U S^{\frac{1}{2}} \Sigma(\bm{x}) (US^{\frac{1}{2}})^T, \label{vyx}
\end{align}
where $\sigma^2$ denotes noise in the data and constitutes a hyperparameter. Eq. \ref{eq:mean} and Eq. \ref{vyx} follow directly from Eq. \ref{pyz} and Eq. \ref{pzx} and associate coordinates of each vertex in the predicted surface mesh with an explicitly expressed uncertainty.  Full derivation can be found in Appendix A.

\subsubsection{Training}
In order to predict $\Sigma(\bm{x})$, $\mu(\bm{x})$ and $s(\bm{x})$, we train a deep CNN by minimising an objective function $L$ consisting of two components: a surface prediction fidelity data term $L_{data}$ aiming to maximise the conditional probability $p(y|\bm{x})$, and a regularisation term $L_{reg}$ minimising the distance between the prior and observed distribution over $z$ by means of  Kullback-Leibler divergence (KLD)
\begin{equation}
    L = \lambda \, L_{reg}- L_{data},
\end{equation}
where $\lambda >0$ is a regularisation parameter.

Whilst Eq. \ref{pyx} allows for direct maximisation of $\ln p(y|\bm{x})$ by marginalisation of the latent variable, this approach is not computationally feasible, as outlined in \cite{workshop}. Even though the marginal distribution is Gaussian with closed form mean and variance, direct optimisation would require inversions of a large and often poorly conditioned covariance matrix (\ref{vyx}) which could lead to numerical instabilities as was empirically observed.
We apply Jensen's inequality to Eq. \ref{pyx} and derive a lower bound for $\ln p(y|\bm{x})$, which will constitute our prediction fidelity data term as follows
\begin{align}
    \ln p(y|\bm{x}) \geq \mathbb{E}_{z|\bm{x}}\left[ \ln p(y|z,\bm{x}) \right] 
                                                \cong \frac{1}{L} \sum_{l=1}^L \ln p(y|z_l, \bm{x}), \label{post}
\end{align}
where $z_l$ is sampled from $p(z|\bm{x})$ defined in Eq. \ref{pzx}. Sampling is done by means of a ``reparametrisation trick'' \cite{kingma} through $z_l = \mu(x) + \Sigma^{1/2}_x * \epsilon$, $\epsilon \sim \mathcal{N}(0,I)$. $L$ refers to the number of samples, in our case $L=1$. Hence 
\begin{equation}
    L_{data} = \sum_{n=1}^N \frac{1}{L} \sum_{l=1}^L \ln p(y|z_l, \bm{x_n}).
\end{equation}

Maximisation of the lower bound in Eq.~\ref{post} might not satisfy the prior set on the latent variable $z$ in our probabilistic PCA model ($z\sim\mathcal{N}(0,I)$). To align the distribution of $z$ observed in the training data $p(z) \cong \sum p(z|\bm{x}_n)$ obtained from the input image stack $ \bm{x}_n \sim p(\bm{x})$  with the prior, we employ the KLD regularisation and minimise 
\begin{equation}\label{KLD}
	L_{reg} = \mathrm{KLD} \left( \sum_{n=1}^{N} p(z|\bm{x}_n), p(z) \right).
\end{equation}

For further details on the derivations please refer to Appendix B.

\subsubsection{Spatially Aware Deep Network Architecture}
To predict the distribution of the latent space along with the global shift of the surface, we build a 3-branch deep CNN where each branch takes one MR imaging slice as input, see Fig. \ref{fig:framework}. In each branch, 9 convolutional layers (stride = 1) are intertwined with 3 max pooling layers (stride = 2). Last features of the convolutional branches are concatenated and passed to two dense layers. All convolutional and the first dense layers are followed by ReLU activations. 

An important challenge to be addressed is the fact that all three available input slices pertaining to a given organ are related in the real 3D world. Each slice is associated with a separate transformation matrix between the pixel and world coordinate system. The reference and reconstruction meshes are both defined in absolute world space coordinates. Hence, working solely in pixel space and applying the CNN to 2D images directly would lead to inconsistencies and poor performance as the accurate spatial localisation and orientation of image slices with respect to each other as well as to the reference meshes would be lost. Inspired by \cite{coordconv}, we propose a simple yet effective solution to mitigate this issue by taking advantage of the known transformation matrices between pixel and world coordinates. These are available for every MRI volume. More precisely, the input to every network branch will consist of intensities of each MR imaging slice $\bm{X}_i$ ($i=1,2,3$) concatenated with the coordinates of each pixel in the 3D world space. 

Alternative complex solutions such as spatial transformers \cite{spatialTransformer} are infeasible in this situation as they would require converting the input 2D imaging plane into positions in 3D space and hence a construction of a full imaging 3D volume. This is however impracticable due to missing imaging information outside of the original imaging plane, since all we have are 2D slices. 

\section{Results}
\begin{table}[t]
\caption{Results of surface prediction: average statistics collected over the test set. Results of our method were computed using mean predictions. Average DICE$_{pair}$ score were computed over all pairs of predicted shapes across all subjects in the aligned test set and reflects an inter-subject shape variability.}\label{tab:coordconv}
\begin{center}
\begin{tabular}{|l|l|l|l|l||l|}
\hline
Method & DICE & HD & ASSD & RAVD & DICE$_{pair}$ \\
\hline

det PCA$_{O}$ 12 & 0.15 $\pm$ 0.12 & 19.1 $\pm$ 10.3 & 7.38 $\pm$ 5.40 & -15.6 $\%$& \multicolumn{1}{c|}{-} \\ 
det PCA$_{O}$ 16 & 0.14 $\pm$ 0.11 & 19.7 $\pm$ 10.4 & 7.50 $\pm$ 5.52 & -14.6 \%& \multicolumn{1}{c|}{-}\\ 
det PCA$_{O}$ 20 & 0.15 $\pm$ 0.11 & 19.2 $\pm$ 10.0 & 7.35 $\pm$ 5.31 & -12.6 \%& \multicolumn{1}{c|}{-}\\ 

det PCA 12 & 0.57 $\pm$ 0.15 & 5.04 $\pm$ 1.90 & \textbf{1.02 $\pm$ 0.54} & 6.09 \%& \multicolumn{1}{c|}{-}\\ 
det PCA 16 & 0.56 $\pm$ 0.15 & 5.43 $\pm$ 2.38 & 1.08 $\pm$ 0.70 & 3.12 \%& \multicolumn{1}{c|}{-}\\ 
det PCA 20 & 0.58 $\pm$ 0.14 & 5.09 $\pm$ 1.87 & 1.02 $\pm$ 0.58 & 0.97 \%& 0.49 $\pm$ 0.21\\

TL-net & 0.59 $\pm$ 0.19 & 7.75 $\pm$ 6.55 & 1.37 $\pm$ 0.76 & -23.8 \%& 0.52 $\pm$ 0.22 \\ 
REC-CVAE & \textbf{0.61 $\pm$ 0.13} & 8.02 $\pm$ 1.24 & 1.18 $\pm$ 0.31 & -5.62 \%&  0.97 $\pm$ 0.02\\\hline
\hline
Ours prob PCA 12 & 0.54 $\pm$ 0.15 & 5.60 $\pm$ 1.92 & 1.15 $\pm$ 0.62 & 0.44 \% & \multicolumn{1}{c|}{-}\\%
Ours prob PCA 16 & 0.55 $\pm$ 0.13 & 5.37 $\pm$ 2.02 & 1.06 $\pm$ 0.52 & 0.63 \% & \multicolumn{1}{c|}{-}\\ %
Ours prob PCA 20 & 0.57 $\pm$ 0.13 & \textbf{4.98 $\pm$ 1.66} & \textbf{1.02 $\pm$ 0.54} & \textbf{0.29 $\%$} & 0.70 $\pm$ 0.20\\ %
\hline
\end{tabular}
\end{center}
\end{table} 
The proposed method was evaluated on the task of the left ventricle myocardium surface reconstruction using cardiac imaging volumes from UK BioBank \cite{UKBB}. We predict 3D coordinates of 22043 surface mesh vertices from 3 quasi-orthogonal MR image slices: one acquired along the short axis and 2 along the long axis. All images were resampled to isotropic pixel size of 1.8269\,mm and cropped / padded to the size of $80 \times 80$ pixels. The data set used consists of 529 training, 178 testing and 178 validation examples. Reference meshes used for training and evaluation were constructed using an atlas based method described in \cite{puyol2017} from the segmentations prepared automatically using expert-segmentations and combination of learning and registration methods from \cite{bai2017b,sinclair2017}. Mesh connectivity was fixed throughout the data set. Training was done via an RMS-Prop optimiser with a constant learning rate of $10^{-6}$ for batches of size 5. Hyperparameters were optimised on the validation set. Noise level in the data $\sigma^2$ and regularisation parameter $\lambda$ were empirically set to $\sigma^2 = 5\times 10^{-2}$ and $\lambda = 10^{3}$ respectively.

\begin{figure}[t]
    \centering
    \includegraphics[scale=0.06]{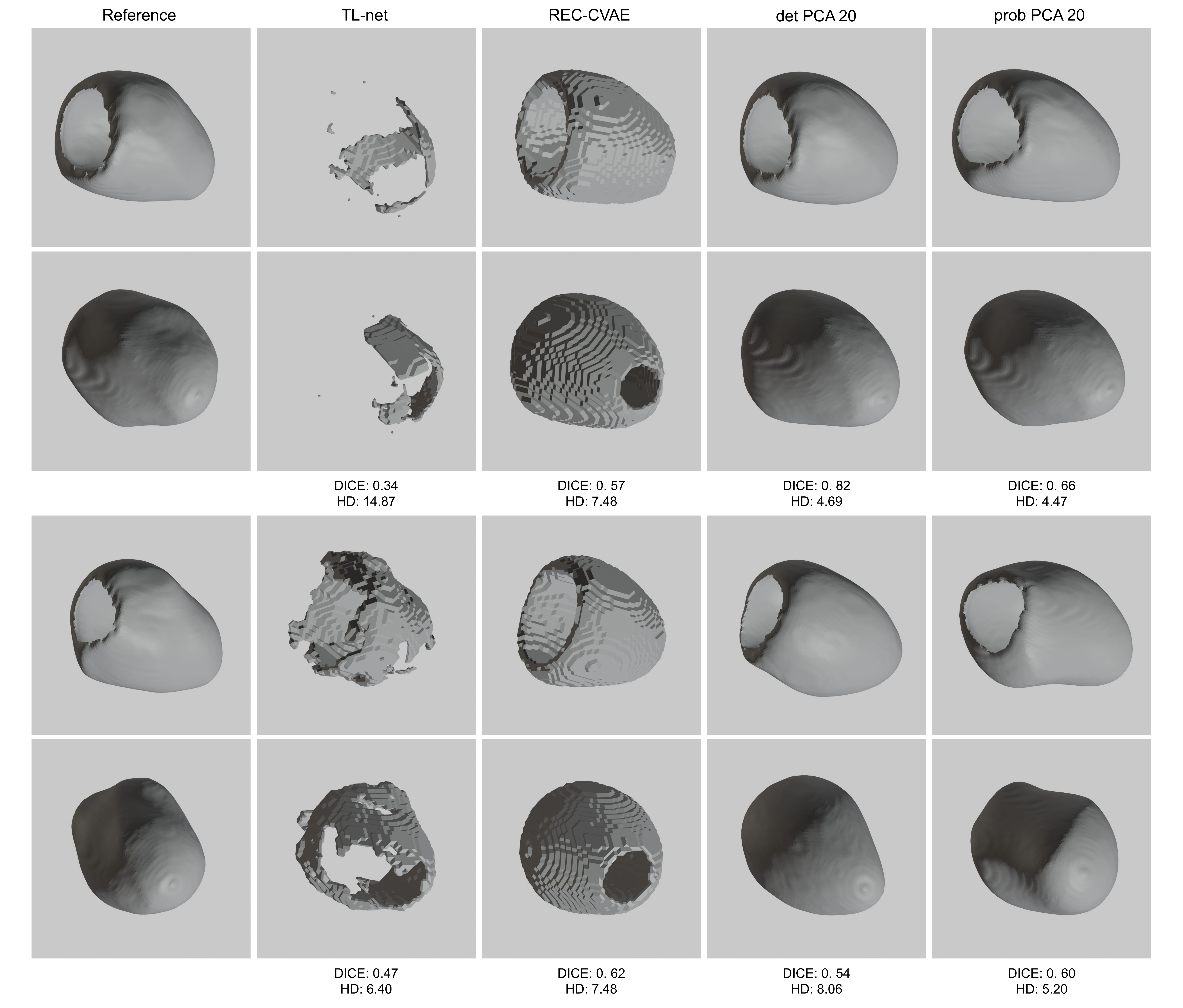}
    \caption{Example of shape predictions for two different subjects visualised from two distinct angles.}
    \label{fig:predictions}
\end{figure}
 We compared our method (prob PCA $n$, $n$ = number of principal components used) with four baselines: 1) deterministic PCA  (det PCA$_O$ $n$) as proposed in \cite{milletari2017} without the spatial transformer refinement and extended to 3D using an architecture analogous to ours, 2) det PCA enriched with input coordinate maps as in our method, 3) TL-net architecture \cite{TL-net} implemented as in \cite{juan2018} and 4) conditional variational autoencoder method REC-CVAE from \cite{juan2018}. The last two methods were trained and evaluated on a pre-aligned set of reference volumes.

Quantitative results can be seen in Table \ref{tab:coordconv}. The models were evaluated using a mixture of volumetric and surface measures: DICE score, Hausdorff distance (HD), Average symmetric surface distance (ASSD), and Relative absolute volume difference (RAVD). Where applicable, volumetric measures were computed using voxelisations of reference and predicted meshes. Surface measures were then aggregated on surfaces extracted from volumetric shapes. Results of our probabilistic model were evaluated using the mean prediction defined in Eq. \ref{eq:mean}. 
We can see that our approach outperforms all competing methods in terms of surface criteria and RAVD. Fig. \ref{fig:predictions} shows that our results are qualitatively superior to coarse volumetric shapes of unrestricted topology obtained from volumetric methods. Our predicted meshes are genus 0 and respect the volume of the reconstructed organ. 

\begin{figure*}[t]
\begin{center}
\begin{minipage}[c]{0.8\linewidth}
\begin{center}
\begin{tabular}[t]{cc}
\begin{minipage}[c]{0.8\linewidth}
\begin{minipage}[c]{0.19\linewidth}
	\includegraphics[angle=0,origin=c, width=\linewidth]{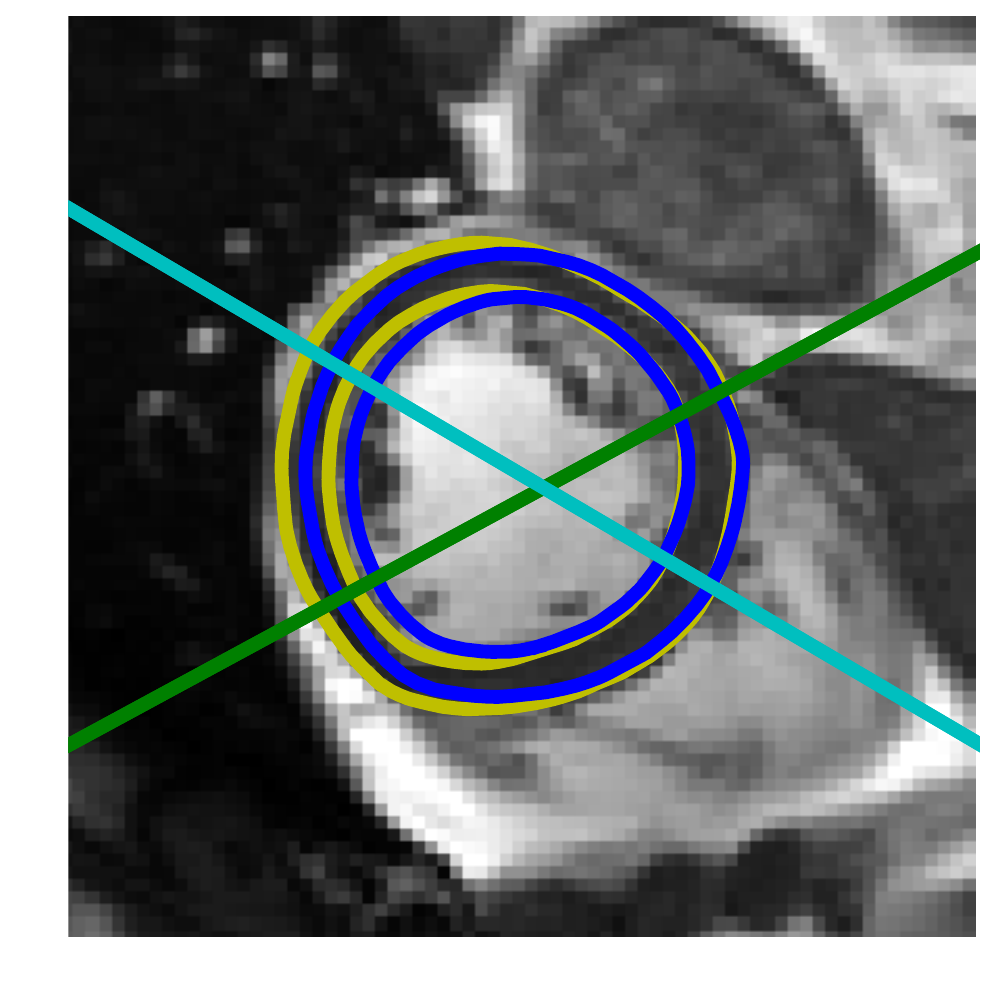}
\end{minipage}
\begin{minipage}[c]{0.19\linewidth}
	\includegraphics[angle=0,origin=c, width=\linewidth]{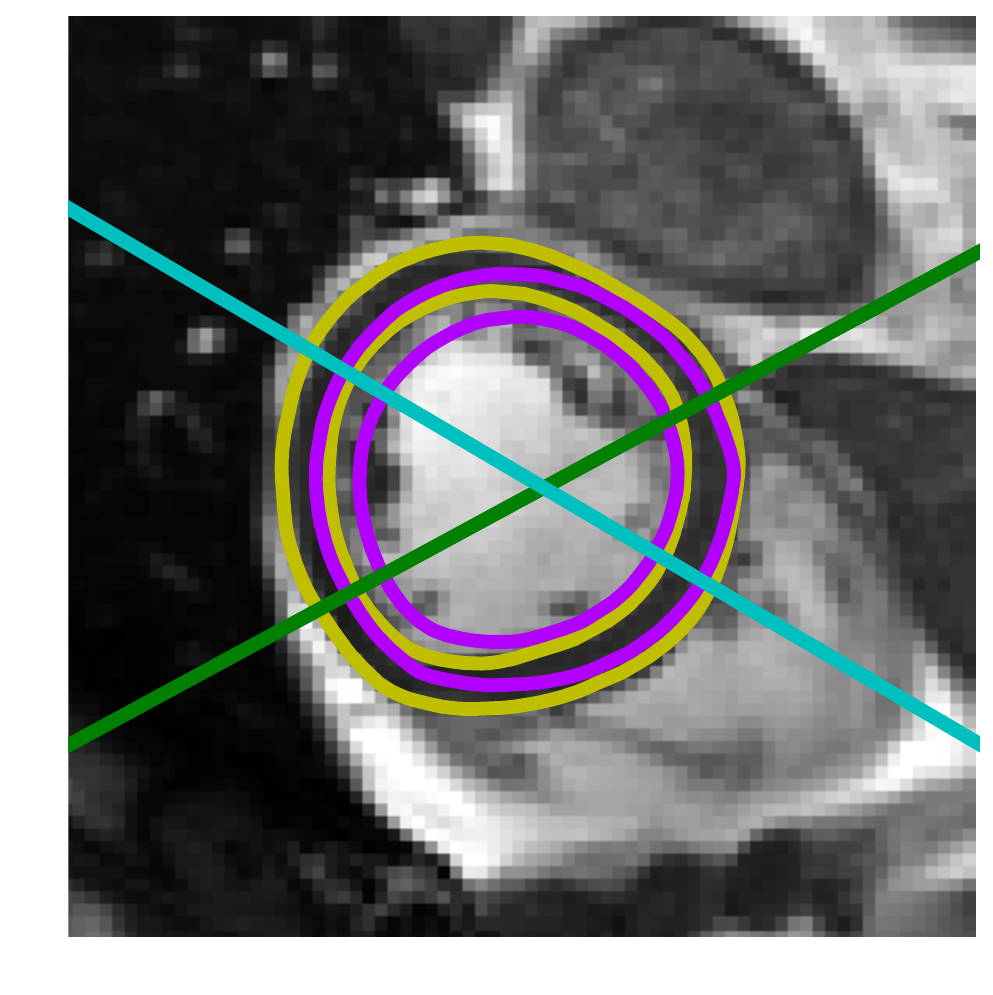}
\end{minipage}
\begin{minipage}[c]{0.19\linewidth}
	\includegraphics[angle=0,origin=c, width=\linewidth]{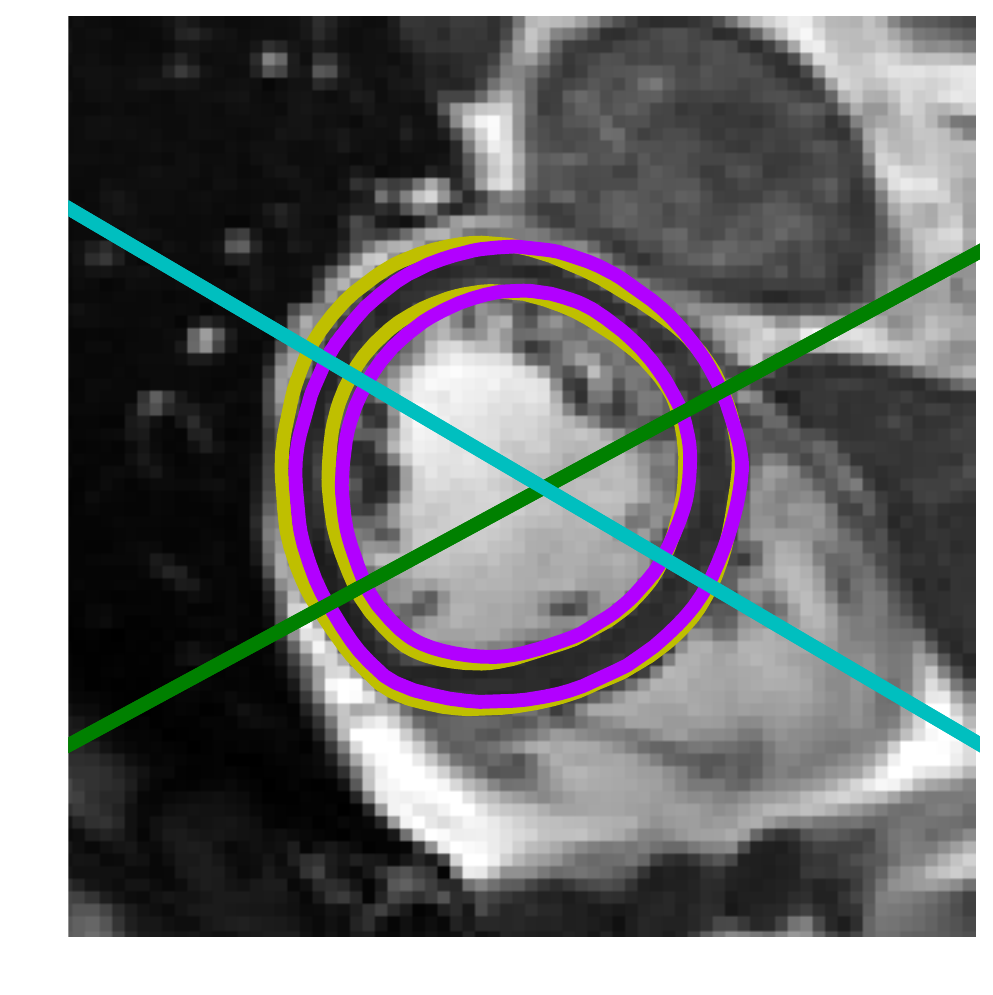}
\end{minipage}
\begin{minipage}[c]{0.19\linewidth}
	\includegraphics[angle=0,origin=c, width=\linewidth]{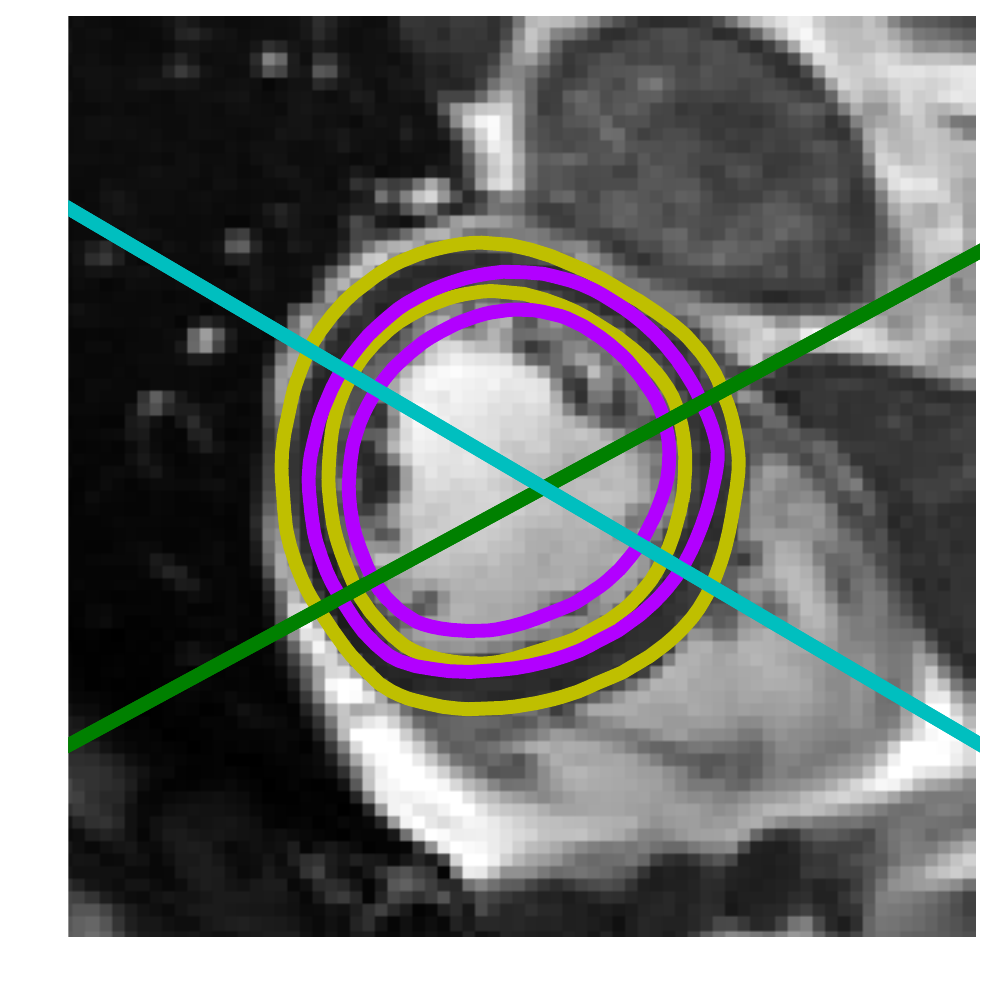}
\end{minipage}
\begin{minipage}[c]{0.19\linewidth}
	\includegraphics[angle=0,origin=c, width=\linewidth]{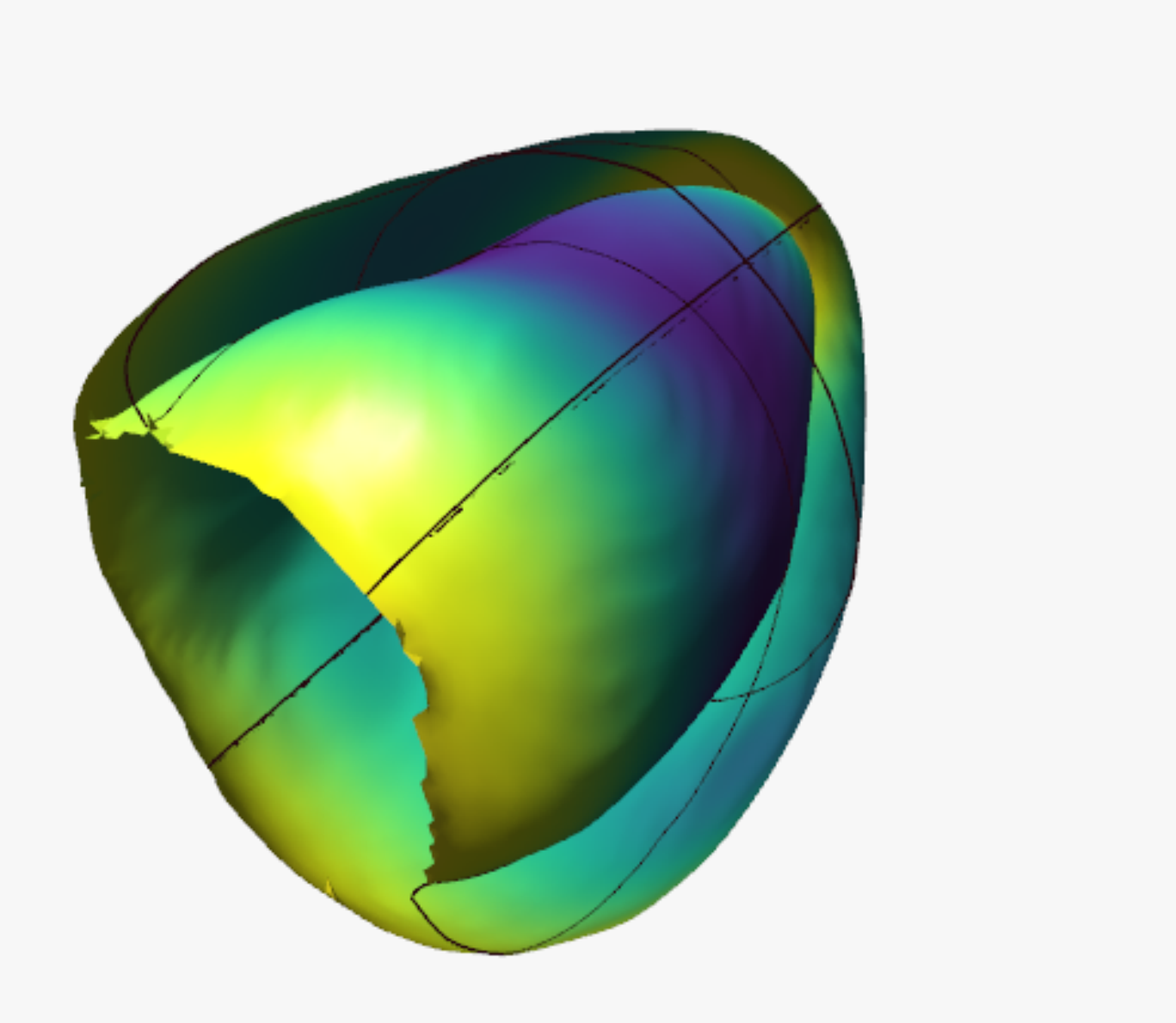}
\end{minipage}
\begin{minipage}[c]{0.19\linewidth}
	\includegraphics[angle=0,origin=c, width=\linewidth]{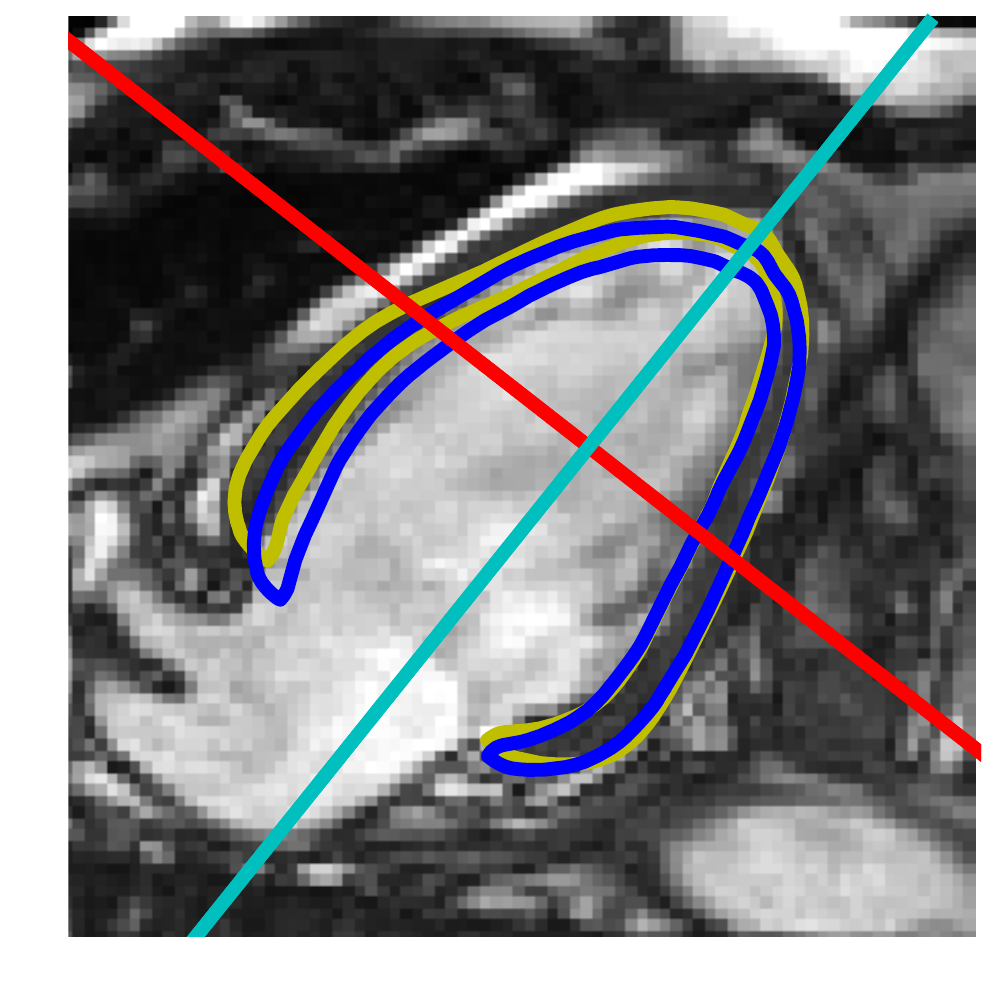}
\end{minipage}
\begin{minipage}[c]{0.19\linewidth}
	\includegraphics[angle=0,origin=c, width=\linewidth]{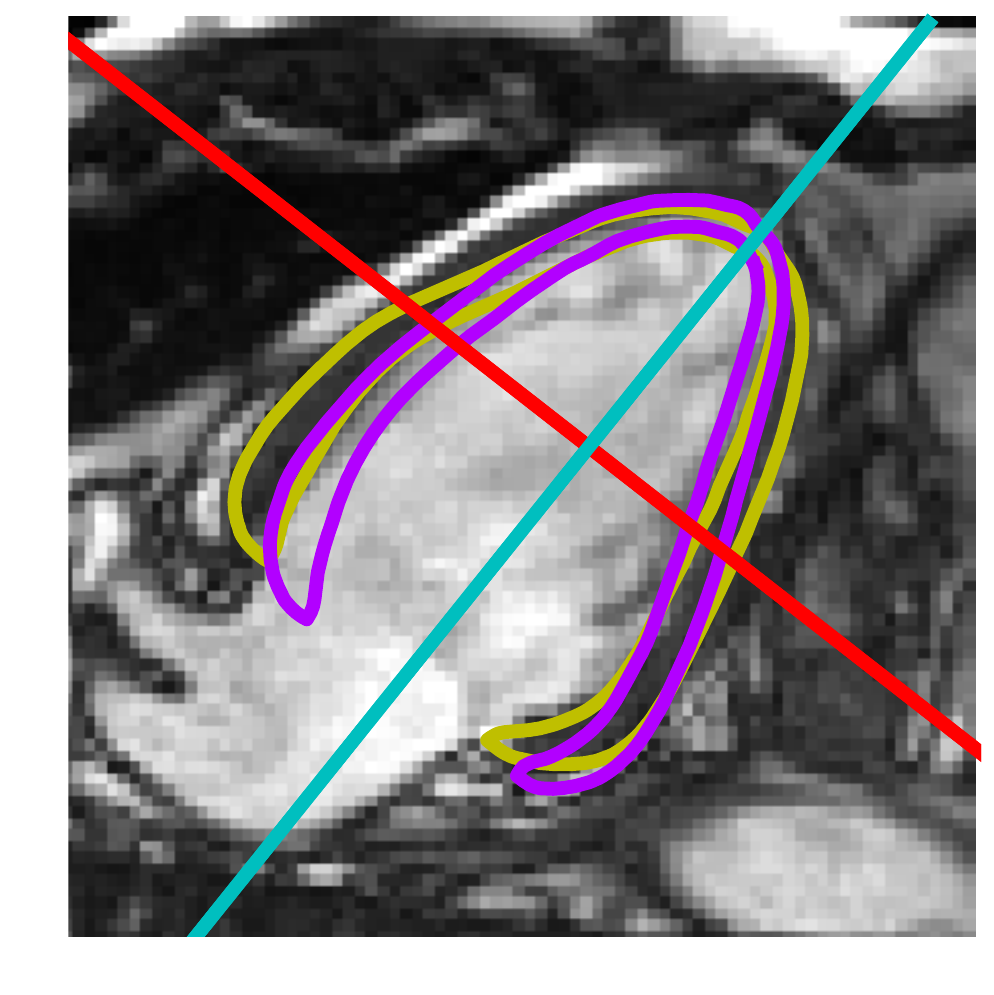}
\end{minipage}
\begin{minipage}[c]{0.19\linewidth}
	\includegraphics[angle=0,origin=c, width=\linewidth]{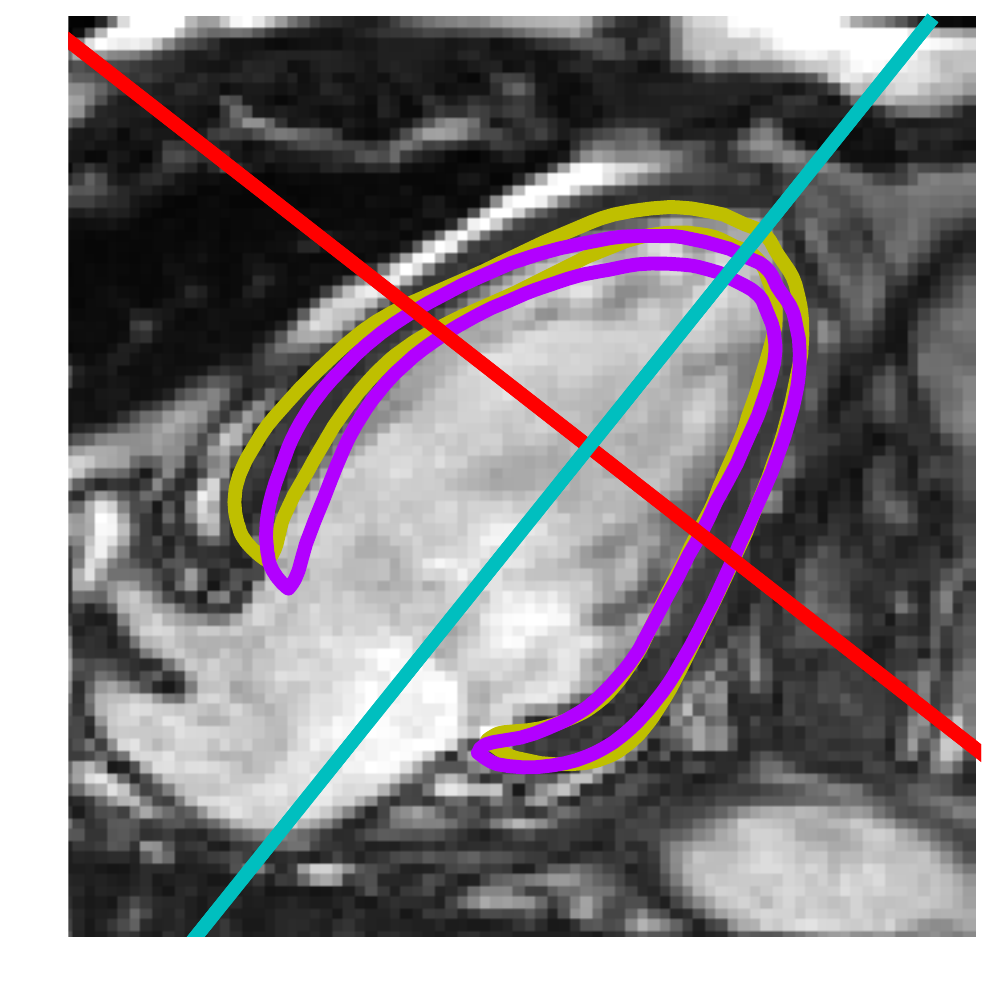}
\end{minipage}
\begin{minipage}[c]{0.19\linewidth}
	\includegraphics[angle=0,origin=c, width=\linewidth]{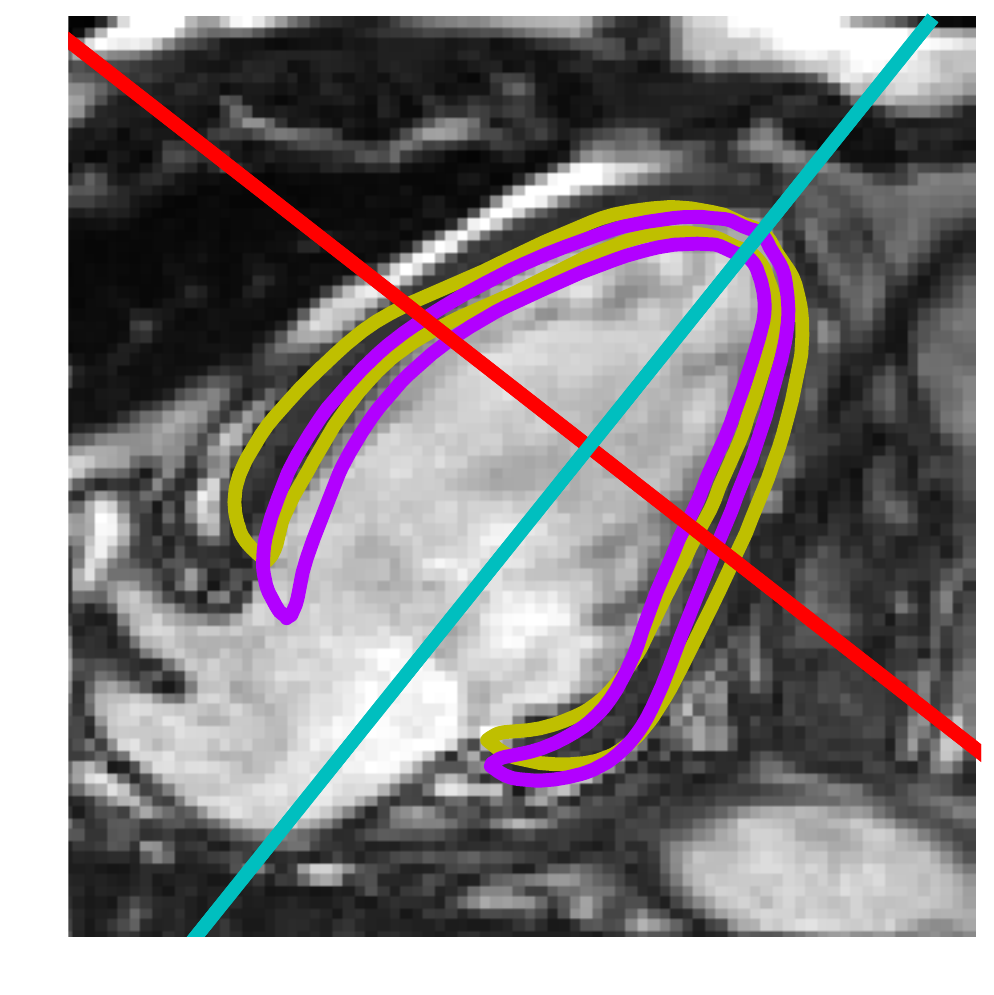}
\end{minipage}
\begin{minipage}[c]{0.19\linewidth}
	\includegraphics[angle=0,origin=c, width=\linewidth]{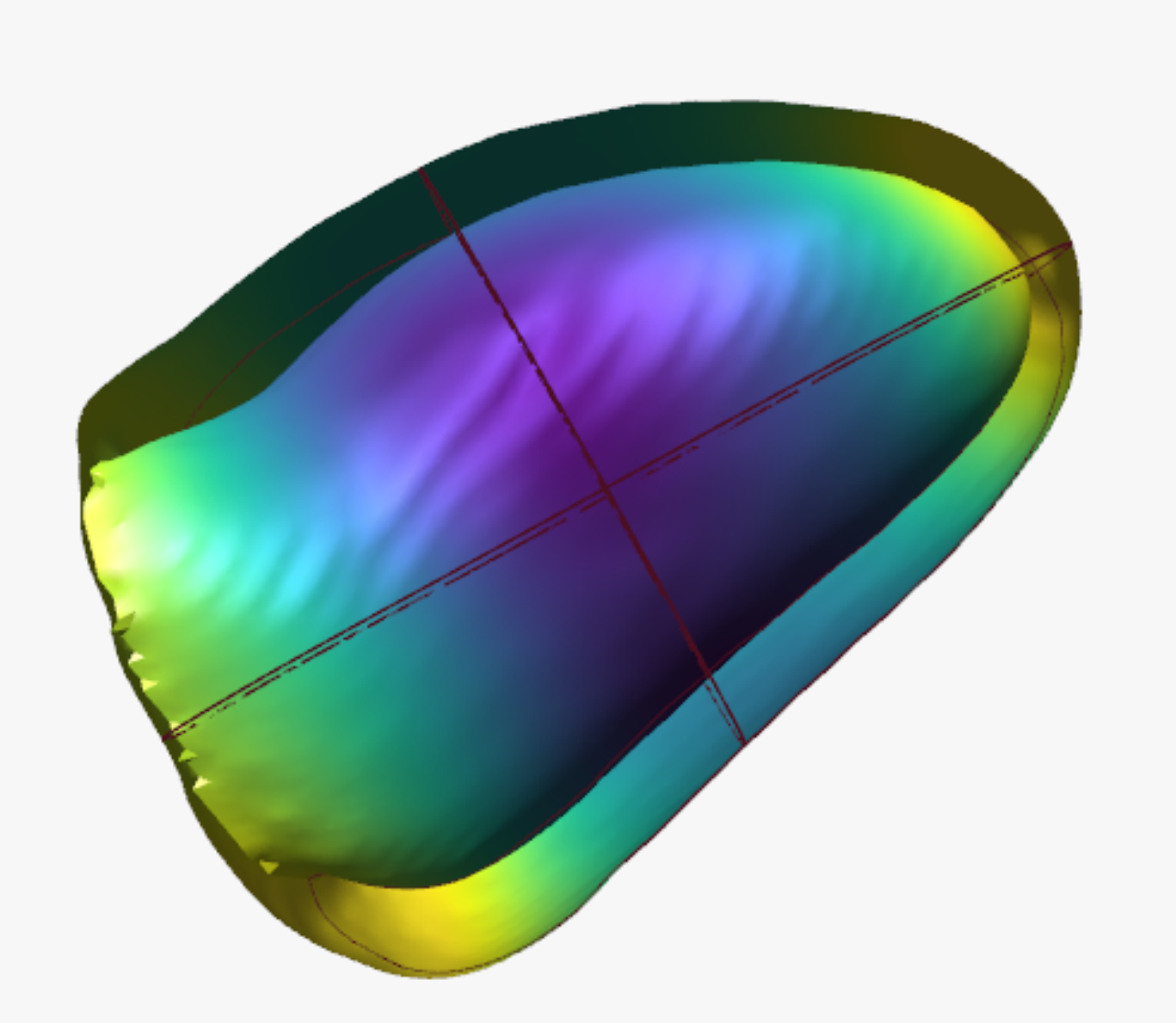}
\end{minipage}
\begin{minipage}[c]{0.19\linewidth}
	\includegraphics[angle=0,origin=c, width=\linewidth]{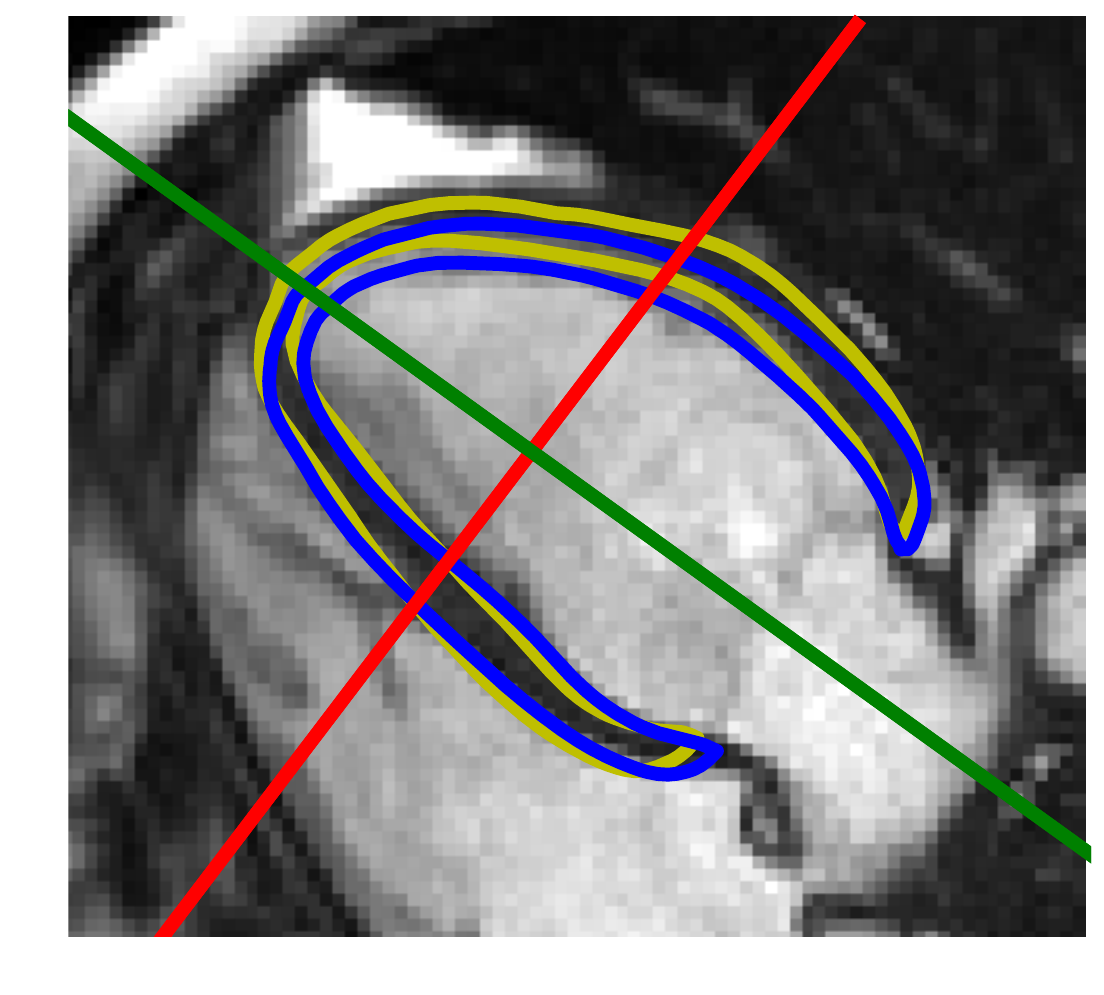}
\end{minipage}
\begin{minipage}[c]{0.19\linewidth}
	\includegraphics[angle=0,origin=c, width=\linewidth]{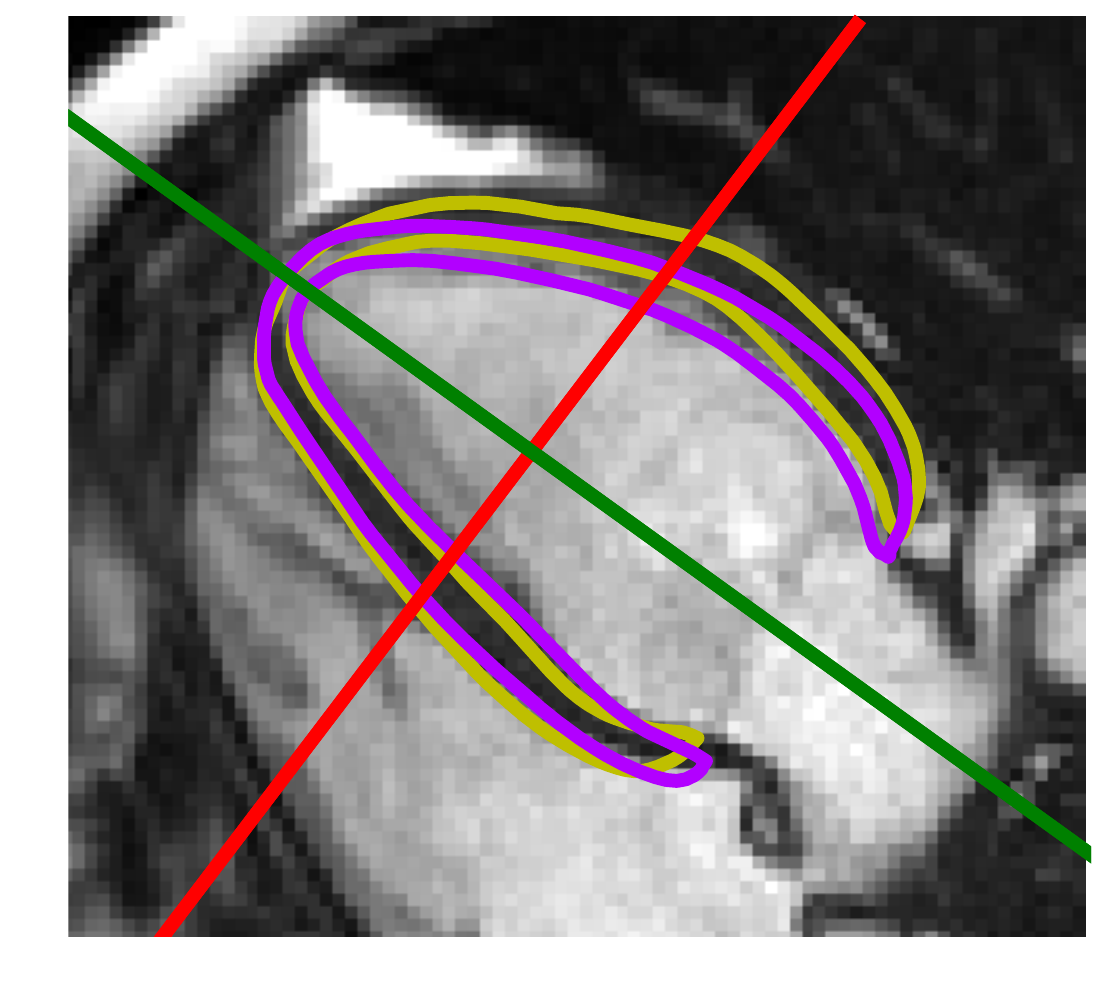}
\end{minipage}
\begin{minipage}[c]{0.19\linewidth}
	\includegraphics[angle=0,origin=c, width=\linewidth]{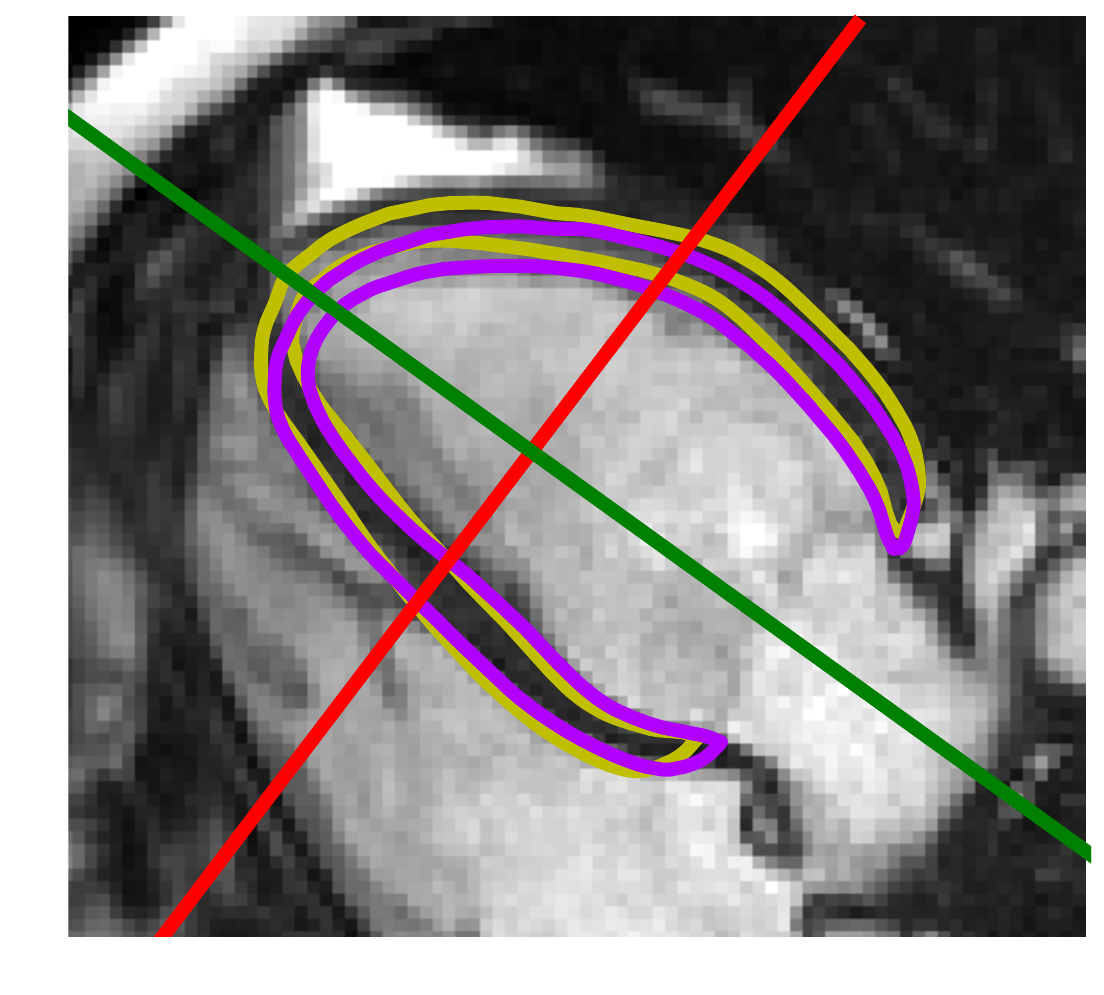}
\end{minipage}
\begin{minipage}[c]{0.19\linewidth}
	\includegraphics[angle=0,origin=c, width=\linewidth]{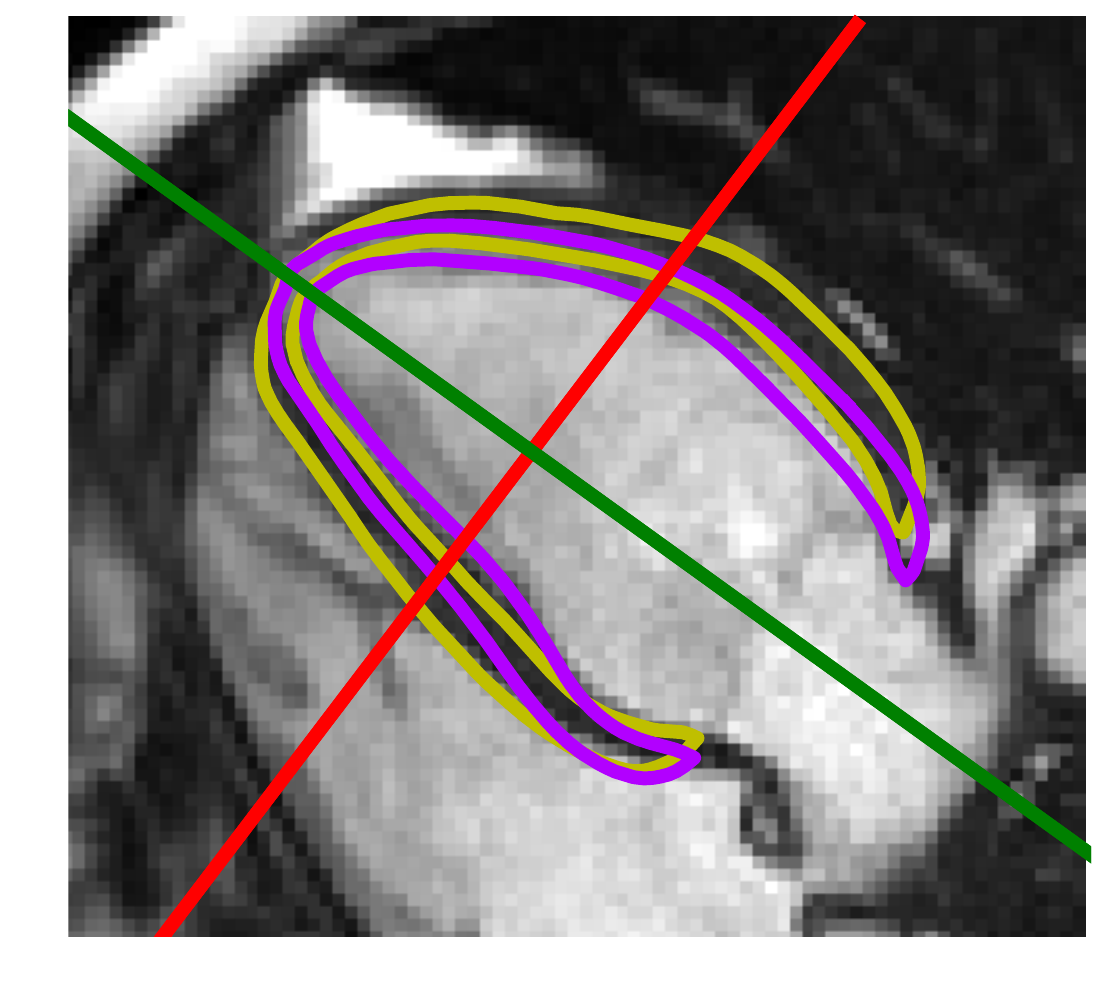}
\end{minipage}
\begin{minipage}[c]{0.19\linewidth}
	\includegraphics[angle=0,origin=c, width=\linewidth]{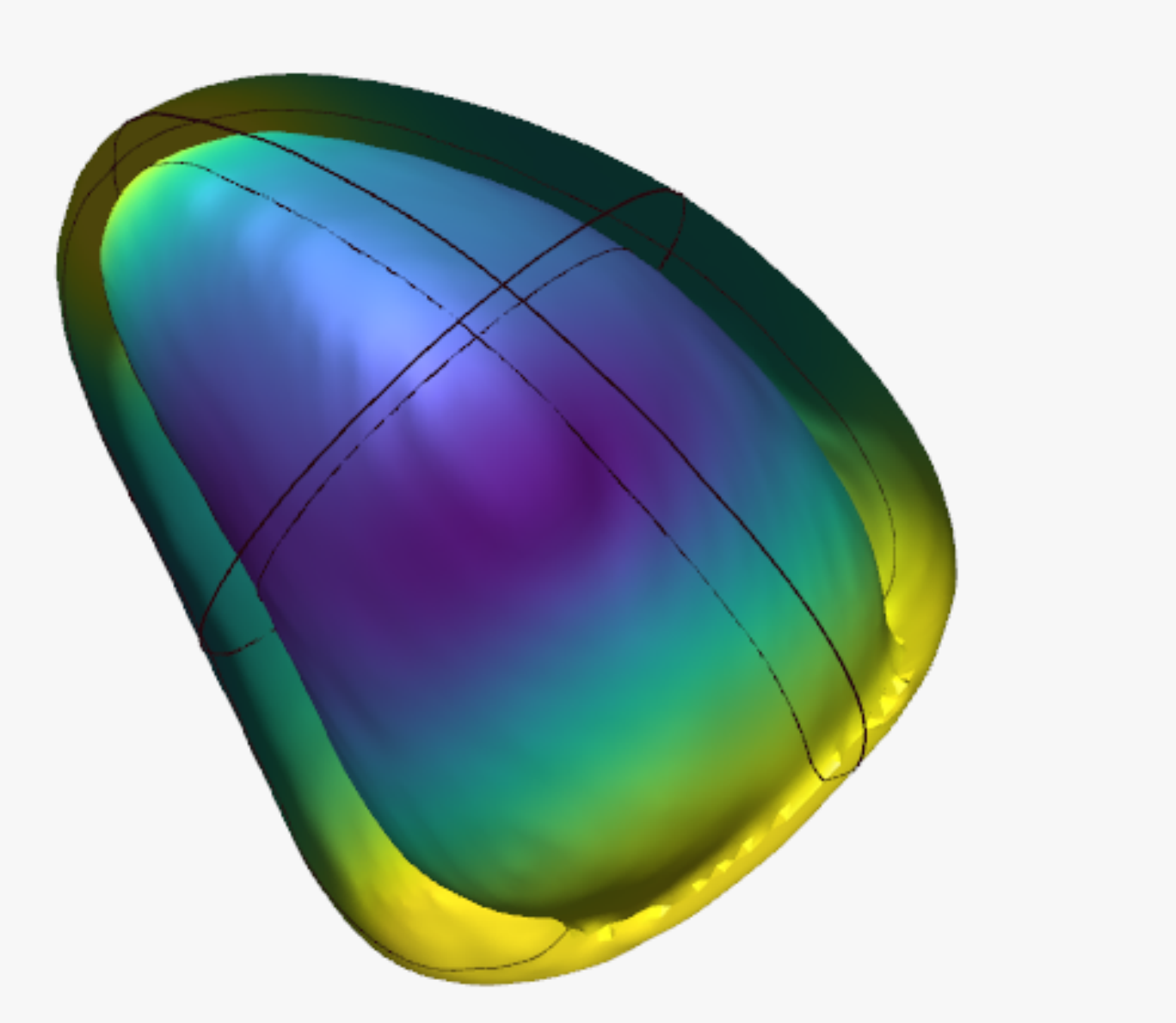}
\end{minipage}
\end{minipage}
&
\begin{minipage}[c]{0.05\linewidth}
	\includegraphics[angle=0,origin=c, width=0.94\linewidth]{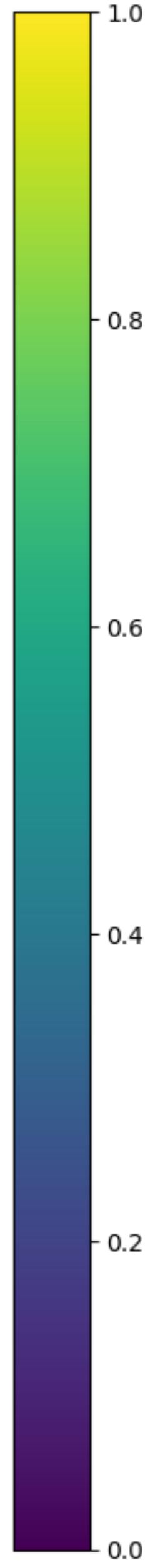}
\end{minipage}
\end{tabular}
\end{center}
\end{minipage}
\end{center}
\caption{Surface reconstruction results for a single subject. Column 1 on the left: intersection of the 3D surface with the input MR imaging slices - ground truth surface (yellow), mean surface prediction (blue). Straight lines represent the intersection of the other two input MR slices with the current imaging plane. Columns 2 - 4: Example surface samples (magenta). Column 5: Prediction uncertainty visualisation - colour of the heatmap at each vertex of the mean shape mesh corresponds to the $\log \mathrm{det} \Sigma_i$, $\Sigma_i \in \mathbb{R}^{3\times 3}$ is the covariance matrix at the given vertex, rescaled onto interval $[0,1]$. \label{glyphs}} 
\end{figure*}

We observe poorer performance in terms of DICE score, which can be explained through its definition. 
Methods that predict empty space in uncertain areas garner superior scores as they minimise the number of false positives in the prediction. 
We further quantified predicted shape variability for the top performing methods by means of an average DICE$_{pair}$ score computed over all pairs of predicted shapes across all subjects in the test set---the higher the DICE, the lower the shape variability. Shapes were aligned first: in REC-CVAE \cite{juan2018} and TL-net \cite{TL-net} by design of the prediction process, in det PCA \cite{milletari2017} and our method were aligned ex post using PCA. The high DICE$_{pair}$ of 0.97 in REC-CVAE suggests the method fails to diversify the predicted shapes. For reference, the ground truth meshes have a DICE$_{pair}$ equal to $0.42  \pm 0.17$. Fig.~\ref{glyphs} exemplifies the probabilistic nature of our method by visualising the mean surface prediction and sampled surface meshes by means of their intersections with the input imaging slices. Prediction uncertainty at surface vertices can be assessed via a heatmap - vertex colours correspond to scaled $\log \mathrm{det} \Sigma_i$, where $\Sigma_i \in \mathbb{R}^{3\times 3}$ is the covariance matrix at the given vertex extracted from $\mathrm{var}(y|\bm{x})$. The larger the variance, the bigger the uncertainty over the position of the vertex that can be sampled from this distribution. Notice how uncertainty grows with the increasing distance from the center of the organ. Additional qualitative examples are available in Appendix C.

\section{Conclusion}
In this paper, we presented a novel probabilistic deep learning approach for concurrent 3D surface reconstruction from sparse 2D MR image data and aleatoric uncertainty prediction. Our method is capable of reconstructing high resolution large surface meshes from three quasi-orthogonal MR imaging slices from limited training data whilst modelling the location of each mesh vertex through a Gaussian distribution. We build on the principles of the probabilistic PCA \cite{bishop} and 2D surface prediction methods from \cite{milletari2017,workshop} and incorporate prior shape information via a linear PCA model. Experiments using cardiac MRI data from UK BioBank \cite{UKBB} show that our method qualitatively and quantitatively outperforms the deterministic and autoencoder baselines in shape reconstruction while correctly localising and orientating the prediction. Moreover, it enables generation of plausible surface reconstructions through sampling from the predicted model and evaluation of the prediction uncertainty. Future work will concentrate on generalisation of the method to surface reconstruction in the presence of pathologies and reconstruction of other types of organs of more variable shapes, which may lead to adaptation of the used shape model.

\subsubsection{Acknowledgements}
This research has been conducted using the UK Biobank Resource under Application Number 17806.

\bibliographystyle{splncs04}
\bibliography{bibliography}

\begin{thebibliography}{10}
\providecommand{\url}[1]{\texttt{#1}}
\providecommand{\urlprefix}{URL }
\providecommand{\doi}[1]{https://doi.org/#1}

\bibitem{bai2017b}
Bai, W., Oktay, O., Sinclair, M., Suzuki, H., Rajchl, M., Tarroni, G., et~al.:
  Semi-supervised learning for network-based cardiac {MR} image segmentation.
  In: Descoteaux, M., Maier{-}Hein, L., et~al. (eds.) Medical Image Computing
  and Computer Assisted Intervention - {MICCAI} 2017 - 20th International
  Conference, Quebec City, QC, Canada, September 11-13, 2017, Proceedings, Part
  {II}. Lecture Notes in Computer Science, vol. 10434, pp. 253--260. Springer
  (2017)

\bibitem{bai2013}
Bai, W., Shi, W., O'Regan, D.P., Tong, T., Wang, H., Jamil{-}Copley, S.,
  et~al.: A probabilistic patch-based label fusion model for multi-atlas
  segmentation with registration refinement: Application to cardiac {MR}
  images. {IEEE} Trans. Medical Imaging  \textbf{32}(7),  1302--1315 (2013)

\bibitem{bai2018}
Bai, W., Sinclair, M., Tarroni, G., Oktay, O., Rajchl, M., Vaillant, G.,
  et~al.: Automated cardiovascular magnetic resonance image analysis with fully
  convolutional networks. Journal of Cardiovascular Magnetic Resonance
  \textbf{20} (12 2018)

\bibitem{baumgartner2019}
Baumgartner, C.F., Tezcan, K.C., Chaitanya, K., H{\"{o}}tker, A.M.,
  Muehlematter, U.J., Schawkat, K., et~al.: Phiseg: Capturing uncertainty in
  medical image segmentation. In: Shen, D., Liu, T., et~al. (eds.) Medical
  Image Computing and Computer Assisted Intervention - {MICCAI} 2019 - 22nd
  International Conference, Shenzhen, China, October 13-17, 2019, Proceedings,
  Part {II}. Lecture Notes in Computer Science, vol. 11765, pp. 119--127.
  Springer (2019)

\bibitem{bhalodia2018}
Bhalodia, R., Elhabian, S.Y., Kavan, L., Whitaker, R.T.: Deepssm: {A} deep
  learning framework for statistical shape modeling from raw images. In:
  Reuter, M., Wachinger, C., et~al. (eds.) Shape in Medical Imaging -
  International Workshop, ShapeMI 2018, Held in Conjunction with {MICCAI} 2018,
  Granada, Spain, September 20, 2018, Proceedings. Lecture Notes in Computer
  Science, vol. 11167, pp. 244--257. Springer (2018)

\bibitem{bishop}
Bishop, C.M.: Pattern recognition and machine learning, 5th Edition.
  Information science and statistics, Springer (2007)

\bibitem{juan2018}
Cerrolaza, J.J., Li, Y., Biffi, C., G{\'{o}}mez, A., Sinclair, M., Matthew, J.,
  et~al.: 3d fetal skull reconstruction from 2dus via deep conditional
  generative networks. In: Frangi, A.F., Schnabel, J.A., et~al. (eds.) Medical
  Image Computing and Computer Assisted Intervention - {MICCAI} 2018 - 21st
  International Conference, Granada, Spain, September 16-20, 2018, Proceedings,
  Part {I}. Lecture Notes in Computer Science, vol. 11070, pp. 383--391.
  Springer (2018)

\bibitem{chang2011}
Chang, J., III, J.W.F.: Efficient {MCMC} sampling with implicit shape
  representations. In: The 24th {IEEE} Conference on Computer Vision and
  Pattern Recognition, {CVPR} 2011, Colorado Springs, CO, USA, 20-25 June 2011.
  pp. 2081--2088. {IEEE} Computer Society (2011)

\bibitem{fischl2012}
Fischl, B.: Freesurfer. NeuroImage  \textbf{62}(2),  774--781 (2012)

\bibitem{garcia-barnes2010}
Garcia{-}Barnes, J., Gil, D., Badiella, L., Hern{\'{a}}ndez{-}Sabat{\'{e}}, A.,
  Carreras, F., Pujadas, S., Mart{\'{\i}}, E.: A normalized framework for the
  design of feature spaces assessing the left ventricular function. {IEEE}
  Trans. Medical Imaging  \textbf{29}(3),  733--745 (2010)

\bibitem{TL-net}
Girdhar, R., Fouhey, D.F., Rodriguez, M., Gupta, A.: Learning a predictable and
  generative vector representation for objects. In: Leibe, B., Matas, J., Sebe,
  N., Welling, M. (eds.) Computer Vision - {ECCV} 2016 - 14th European
  Conference, Amsterdam, The Netherlands, October 11-14, 2016, Proceedings,
  Part {VI}. Lecture Notes in Computer Science, vol.~9910, pp. 484--499.
  Springer (2016)

\bibitem{han2004}
Han, X., Pham, D.L., Tosun, D., Rettmann, M.E., Xu, C., Prince, J.L.: {CRUISE:}
  cortical reconstruction using implicit surface evolution. NeuroImage
  \textbf{23}(3),  997--1012 (2004)

\bibitem{huo2016}
Huo, Y., Plassard, A.J., Carass, A., Resnick, S.M., Pham, D.L., Prince, J.L.,
  Landman, B.A.: Consistent cortical reconstruction and multi-atlas brain
  segmentation. NeuroImage  \textbf{138},  197--210 (2016)

\bibitem{spatialTransformer}
Jaderberg, M., Simonyan, K., Zisserman, A., Kavukcuoglu, K.: Spatial
  transformer networks. In: Cortes, C., Lawrence, N.D., et~al. (eds.) Advances
  in Neural Information Processing Systems 28: Annual Conference on Neural
  Information Processing Systems 2015, December 7-12, 2015, Montreal, Quebec,
  Canada. pp. 2017--2025 (2015)

\bibitem{kendall2017}
Kendall, A., Gal, Y.: What uncertainties do we need in bayesian deep learning
  for computer vision? In: Guyon, I., von Luxburg, U., et~al. (eds.) Advances
  in Neural Information Processing Systems 30: Annual Conference on Neural
  Information Processing Systems 2017, 4-9 December 2017, Long Beach, CA,
  {USA}. pp. 5574--5584 (2017)

\bibitem{kingma}
Kingma, D.P., Welling, M.: Auto-encoding variational bayes. In: Bengio, Y.,
  LeCun, Y. (eds.) 2nd International Conference on Learning Representations,
  {ICLR} 2014, Banff, AB, Canada, April 14-16, 2014, Conference Track
  Proceedings (2014)

\bibitem{le2016}
L{\^{e}}, M., Unkelbach, J., Ayache, N., Delingette, H.: Sampling image
  segmentations for uncertainty quantification. Medical Image Anal.
  \textbf{34},  42--51 (2016)

\bibitem{coordconv}
Liu, R., Lehman, J., Molino, P., Such, F.P., Frank, E., Sergeev, A., Yosinski,
  J.: An intriguing failing of convolutional neural networks and the coordconv
  solution. In: Bengio, S., Wallach, H.M., et~al. (eds.) Advances in Neural
  Information Processing Systems 31: Annual Conference on Neural Information
  Processing Systems 2018, NeurIPS 2018, 3-8 December 2018, Montr{\'{e}}al,
  Canada. pp. 9628--9639 (2018)

\bibitem{denis2019}
Madsen, D., Vetter, T., L{\"{u}}thi, M.: Probabilistic surface reconstruction
  with unknown correspondence. In: et~al., H.G. (ed.) Uncertainty for Safe
  Utilization of Machine Learning in Medical Imaging and Clinical Image-Based
  Procedures - First International Workshop, {UNSURE} 2019, and 8th
  International Workshop, {CLIP} 2019, Held in Conjunction with {MICCAI} 2019,
  Shenzhen, China, October 17, 2019, Proceedings. Lecture Notes in Computer
  Science, vol. 11840, pp. 3--11. Springer (2019)

\bibitem{milletari2017}
Milletari, F., Rothberg, A., Jia, J., Sofka, M.: Integrating statistical prior
  knowledge into convolutional neural networks. In: Descoteaux, M.,
  Maier{-}Hein, L., et~al. (eds.) Medical Image Computing and Computer Assisted
  Intervention - {MICCAI} 2017 - 20th International Conference, Quebec City,
  QC, Canada, September 11-13, 2017, Proceedings, Part {I}. Lecture Notes in
  Computer Science, vol. 10433, pp. 161--168. Springer (2017)

\bibitem{peressutti2017}
Peressutti, D., Sinclair, M., Bai, W., Jackson, T., Ruijsink, J., Nordsletten,
  D., et~al.: A framework for combining a motion atlas with non-motion
  information to learn clinically useful biomarkers: Application to cardiac
  resynchronisation therapy response prediction. Medical Image Anal.
  \textbf{35},  669--684 (2017)

\bibitem{puyol2017}
Puyol{-}Ant{\'{o}}n, E., Sinclair, M., Gerber, B., Amzulescu, M.S., Langet, H.,
  Craene, M.D., et~al.: A multimodal spatiotemporal cardiac motion atlas from
  {MR} and ultrasound data. Medical Image Anal.  \textbf{40},  96--110 (2017)

\bibitem{schuh2017}
{Schuh}, A., {Makropoulos}, A., {Wright}, R., {Robinson}, E.C., {Tusor}, N.,
  {Steinweg}, J., et~al.: A deformable model for the reconstruction of the
  neonatal cortex. In: 2017 IEEE 14th International Symposium on Biomedical
  Imaging (ISBI 2017). pp. 800--803 (2017)

\bibitem{UKBB}
Senn, M.: UK BioBank Homepage (Last accessed 24 June 2018),
  \url{https://www.ukbiobank.ac.uk/about-biobank-uk}

\bibitem{sinclair2017}
Sinclair, M., Bai, W., Puyol{-}Ant{\'{o}}n, E., Oktay, O., Rueckert, D., King,
  A.P.: Fully automated segmentation-based respiratory motion correction of
  multiplanar cardiac magnetic resonance images for large-scale datasets. In:
  Descoteaux, M., Maier{-}Hein, L., et~al. (eds.) Medical Image Computing and
  Computer Assisted Intervention - {MICCAI} 2017 - 20th International
  Conference, Quebec City, QC, Canada, September 11-13, 2017, Proceedings, Part
  {II}. Lecture Notes in Computer Science, vol. 10434, pp. 332--340. Springer
  (2017)

\bibitem{tatarchenko}
Tatarchenko, M., Richter, S.R., Ranftl, R., Li, Z., Koltun, V., Brox, T.: What
  do single-view 3d reconstruction networks learn? In: {IEEE} Conference on
  Computer Vision and Pattern Recognition, {CVPR} 2019, Long Beach, CA, USA,
  June 16-20, 2019. pp. 3405--3414. Computer Vision Foundation / {IEEE} (2019)

\bibitem{workshop}
T{\'{o}}thov{\'{a}}, K., Parisot, S., Lee, M.C.H., Puyol{-}Ant{\'{o}}n, E.,
  Koch, L.M., King, A.P., et~al.: Uncertainty quantification in cnn-based
  surface prediction using shape priors. In: Reuter, M., et~al. (eds.) Shape in
  Medical Imaging - International Workshop, ShapeMI 2018, Held in Conjunction
  with {MICCAI} 2018, Granada, Spain, September 20, 2018, Proceedings. Lecture
  Notes in Computer Science, vol. 11167, pp. 300--310. Springer (2018)

\bibitem{3Dprinting}
Vukicevic, M., Mosadegh, B., Min, J.K., Little, S.H.: Cardiac 3d printing and
  its future directions. JACC: Cardiovascular Imaging  \textbf{10}(2),
  171--184 (2017)

\end{thebibliography}


\begin{thebibliography}{8}

\bibitem{bai2013}
Bai, W., et al.: A Probabilistic Patch-Based Label Fusion Model for Multi-Atlas Segmentation With Registration Refinement: Application to Cardiac MR Images.  IEEE Transactions on Medical Imaging \textbf{32}(7), 1302--1315 (2013)

\bibitem{bai2017b}
Bai, W., et al.: Semi-supervised learning for network-based cardiac MR image segmentation. In: Med. Image Comput. Comput. Assist. Interv (MICCAI), 253--260 (2017)


\bibitem{bai2018}
Bai, W., et al.: Automated cardiovascular magnetic resonance image analysis with fully convolutional networks. J Cardiovasc Magn Reson\textbf{20}(65) (2018)

\bibitem{baumgartner2019}
Baumgartner, C., et al.: PHiSeg: Capturing Uncertainty in Medical Image Segmentation. In: Shen, D., et al. (eds.) Medical Image Computing and Computer Assisted Intervention - MICCAI 2019, LCNS, vol 11765, pp. 119--127. Springer, Shenzhen (2019).

\bibitem{bhalodia2018}
Bhalodia, R. et al.: DeepSSM: A Deep Learning Framework for Statistical Shape Modeling from Raw Images. In: Reuter M., Wachinger C., Lombaert H., Paniagua B., Lüthi M., Egger B. (eds) Shape in Medical Imaging, ShapeMI 2018, LCNS, vol 11167. Springer, Cham (2018)

\bibitem{bishop}
Bishop, C.M.: Pattern Recognition and Machine Learning. Springer, Singapore (2006)

\bibitem{juan2018}
Cerrolaza, J. J., et al.: 3D Fetal Skull Reconstruction from 2DUS via Deep Conditional Generative Networks. In: Frangi A., Schnabel J., Davatzikos C., Alberola-López C., Fichtinger G. (eds) Medical Image Computing and Computer Assisted Intervention – MICCAI 2018, LCNS, vol 11070. Springer, Cham (2018)

\bibitem{chang2011}
Chang, J., Fisher III, J.W.: Efficient MCMC Sampling with Implicit Shape Representations. In: 2011 IEEE Conference on Computer Vision and Pattern Recognition (CVPR 2011), pp. 2081--2088. IEEE, (2011)

\bibitem{denker1991}
Denker, J., LeCunn, Y.: Transforming Neural-Net Output Levels to Probability Distributions. In: Advances in Neural Information Processing Systems 3, Citeseer (1991)

\bibitem{fischl2012}
Fischl ,B.: Freesurfer. Neuroimage \textbf{62}, 774--781 (2012)

\bibitem{gal2016}
Gal, Y.: Uncertainty in Deep Learning. University of Cambridge, Cambridge (2016)

\bibitem{garcia-barnes2010}
Garcia-Barnes, J., et al.: A normalized framework for the design of feature spaces assessing the left ventricular function. Med. Imaging IEEE Trans. \textbf{29}(3), 733--745 (2010)

\bibitem{TL-net}
Girdhar, R., Fouhey, D.F., Rodriguez, M., Gupta, A.: Learning a predictable and generative vector representation for objects. In: Leibe, B., Matas, J., Sebe, N., Welling, M. (eds.) ECCV 2016. LNCS, vol. 9910, pp. 484--499. Springer, Cham (2016).

\bibitem{han2004}
Han, X., et al.: CRUISE: Cortical reconstruction using implicit surface evolution. Neuroimage 23, 997--1012 (2004)

\bibitem{huo2016}
Huo, Y. et al.: Consistent Cortical Reconstruction and Multi-Atlas Brain Segmentation. NeuroImage
138, 197--210 (2016)


\bibitem{spatialTransformer}
Jaderberg, M., Simonyan, K., Zisserman, A., Kavukcuoglu, K.: Spatial Transformer Networks. In: Advances in Neural Information Processing Systems 29 (NIPS 2016), (2016)

\bibitem{kihyuk2015}
Kihyuk, S., Lee, H., Yan, X.: Learning Structured Output Representation using Deep Conditional Generative Models, Advances in Neural Information Processing Systems 28 (NIPS 2015), (2015)

\bibitem{kingma}
Kingma, D.P., Welling, M.: Auto-encoding variational bayes. In: ICLR 2014 (2014)


\bibitem{coordconv}
Liu, R., et al.: An Intriguing Failing of Convolutional Neural Networks and the CoordConv Solution. Advances in Neural Information Processing Systems 31 (NIPS 2018), (2018)

\bibitem{le2016}
L\^{e}, M., Unkelbach, J., Ayache, N., Delingette, H.: Sampling Image Segmentations for Uncertainty Quantification. Medical Image Analysis \textbf{34}, 42--51 (2016)


\bibitem{kendall2017}
Kendall, A., Gal, Y.: What Uncertainties Do We Need in Bayesian Deep Learning for Computer Vision. In: Advances in Neural Information Processing Systems 30 (NIPS 2017), pp. 5574--5584. Curran Associates, Long Beach (2017)



\bibitem{denis2019}
Madsen, D., et al.: Probabilistic Surface Reconstruction with Unknown Correspondence. In: Greenspan H. et al. (eds.) Uncertainty for Safe Utilization of Machine Learning in Medical Imaging and Clinical Image-Based Procedures. CLIP 2019, UNSURE 2019, LCNS, vol 11840. Springer, Cham (2019).

\bibitem{mackay1992}
MacKay, D.J.C: A Practical Bayesian Framework for Backpropagation Networks. Neural Computation \textbf{4}(3), 448--472 (1992)

\bibitem{milletari2017}
Milletari, F., Rothberg, A., Jia, J., Sofka, M.: Integrating Statistical Prior Knowledge into Convolutional Neural Networks. In: Descoteaux, M., et al. (eds.) Medical Image Computing and Computer Assisted Intervention (MICCAI 2017). LCNS, vol 10433, pp 161--168. Cham (2017)

\bibitem{neal1995}
Neal, R.M.: Bayesian Learning for Neural Networks. University of Toronto, Toronto (1995)

\bibitem{oktayACNNs}
Oktay, O., et al.: Anatomically Constrained Neural Networks (ACNNs): Application to Cardiac Image Enhancement and Segmentation.  IEEE TRANSACTIONS ON MEDICAL IMAGING \textbf{37} (2), 384--395 (2018)

\bibitem{peressutti2017}
Peressutti, D., et al.: A framework for combining a motion atlas with non-motion information to learn clinically useful biomarkers: Application to cardiac resynchronisation therapy response prediction. Medical Image Analysis \textbf{35}, 669--684 (2017)

\bibitem{puyol2017}
Puyol-Ant\'{o}n, E., et al.: A multimodal spatiotemporal cardiac motion atlas from MR and ultrasound data. Medical Image Analysis \textbf{40}, 96--110 (2017) 

\bibitem{sinclair2017}
Sinclair, M., Bai, W., Puyol-Ant\'{o}n, E., et al.: Fully automated segmentation-based respiratory motion correction of multiplanar cardiac magnetic resonance images for large-scale datasets. In: Descoteaux, M., et al. (eds.) Medical Image Computing and Computer Assisted Intervention (MICCAI 2017). LCNS, (2017)

\bibitem{schuh2017}
Schuh, A., et al.: A Deformable Model for the Reconstruction of the Neaonatal Cortex. In: IEEE 14th International Symposium on Biomedical Imaging (ISBI 2017), pp. 800--803. IEEE, Melbourne (2017)

\bibitem{tatarchenko}
Tatarchenko, M. et al.: What Do Single-view 3D Reconstruction Networks Learn? In: 2019 IEEE Conference on Computer Vision and Pattern Recognition (CVPR 2019), (2019)

\bibitem{Kerem}
Tezcan, K., et al.: MR Image Reconstruction Using Deep Density Priors. In: IEEE TRANSACTIONS ON MEDICAL IMAGING \textbf{38} (7), 1633--1642 (2019)



\bibitem{workshop}
T\'{o}thov\'{a}, K., et al.: Uncertainty Quantification in CNN-Based Surface Prediction Using Shape Priors. In: Reuter M., Wachinger C., Lombaert H., Paniagua B., Lüthi M., Egger B. (eds) Shape in Medical Imaging, ShapeMI 2018, LCNS, vol 11167. Springer, Cham (2018)

\bibitem{UKBB}
UK BioBank Homepage, \url{https://www.ukbiobank.ac.uk/about-biobank-uk}. Last accessed 24 June 2018

\bibitem{VanMolle2019}
Van Molle, P., et al.: Quantifying Uncertainty of Deep Neural Networks in Skin Lesion Classification. In: Greenspan H. et al. (eds.) Uncertainty for Safe Utilization of Machine Learning in Medical Imaging and Clinical Image-Based Procedures. CLIP 2019, UNSURE 2019, LCNS, vol 11840. Springer, Cham (2019).

\bibitem{3Dprinting}
Vukicevic, M., et al.: Cardiac 3D Printing and its Future Directions. JACC: Cardiovascular Imaging \textbf{10}(20), 171--184 (2017)

\bibitem{walker2016}
Walker, J., Doersch, C., Gupta, A., Hebert, M.:  An Uncertain Future: Forecasting from Static Images using Variational Autoencoders. In: 4th European Conference on Computer Vision (ECCV 2016) AAVL Workshop, (2016) 

\end{thebibliography}

\section*{Appendix A}

Given
\begin{equation}\label{pyyz}
    p(y|z,\bm{x}) = \mathcal{N} ( y | U S^{\frac{1}{2}} z + \mu + s(\bm{x}), {\sigma}^2 I), 
\end{equation}
and
\begin{equation}
    p(z|\bm{x}) = p(z|\mu(\bm{x}), \Sigma(\bm{x})),
\end{equation}
the posterior distribution of the vertex coordinates can be derived as
\begin{align*}
    \mathbb{E}(y|\bm{x}) &= \mathbb{E}(\mathbb{E}(y|z)|\bm{x}) \\
                  &= \mathbb{E}(US^{\frac{1}{2}}z+\mu+s|\bm{x}) \\
                  &=\int (US^{\frac{1}{2}}z+\mu+s) p(z|\bm{x}) dz \\
                  &= U S^{\frac{1}{2}} \mathbb{E}(z|\bm{x}) + \mu + s \\
                  &= U S^{\frac{1}{2}} \mu(\bm{x}) + \mu + s, \\
\end{align*}

and 
\begin{align*}
    \mathrm{var}(y|\bm{x}) &= \mathbb{E}(\mathrm{var}(y|z)|\bm{x}) + \mathrm{var}(\mathbb{E}(y|z)|\bm{x}) \\
                    &= \sigma^2I + \mathrm{var}(US^{\frac{1}{2}} z+\mu+s|\bm{x}) \\
                    &= \sigma^2I + \mathrm{var}(US^{\frac{1}{2}} z|\bm{x}) \\
                    &= \sigma^2I + U S^{\frac{1}{2}} \mathrm{var}(z|\bm{x})(US^{\frac{1}{2}})^T \\
                    &= \sigma^2I + U S^{\frac{1}{2}} \Sigma(\bm{x}) (US^{\frac{1}{2}})^T.
\end{align*}

\section*{Appendix B}
\subsection*{Posterior}
From
\begin{equation}
    p(y|z,\bm{x}) = \mathcal{N} ( y | \underbrace{U S^{\frac{1}{2}}z + \mu + s(\bm{x})}_{:= A}, \underbrace{{\sigma}^2 I}_{:=\Sigma}), 
\end{equation}
and Jensen's inequality:
\begin{align}
    \ln p(y|\bm{x}) \geq \int \ln [p(y|z,\bm{x})] p(z|\bm{x}) dz & = \mathbb{E}_{z|\bm{x}}\left[ \ln p(y|z,\bm{x}) \right] \\
                                                & \cong \frac{1}{L} \sum_{l=1}^L \ln p(y|z_l,\bm{x}),
\end{align}
where $z_l$ is sampled from $p(z|\bm{x}) = \mathcal{N}(z|\mu(\bm{x}), \Sigma(\bm{x}))$ and $\mu(\bm{x})$, $\Sigma(\bm{x})$ are provided by the network. In detail: 
\begin{equation}
    \frac{1}{L} \sum_{l=1}^L \ln p(y|z_l,\bm{x}) = \frac{1}{L} \sum_{l=1}^L -\frac{1}{2} \left\{ r  \ln (2\pi) + \ln |\Sigma| + (y-A)^T \Sigma^{-1}(y-A) \right\},
\end{equation}

where $r$ is the dimensionality of the vector $y$. In practice this value is summed over a batch of input vectors $\bm{x_n}$

\subsection*{Regularisation}
Considering the unitary Gaussian prior on $z$, the fact that
\begin{align}
    p(z) &= \int p(z|\bm{x}) p(\bm{x}) d\bm{x} \\
         & \cong \sum p(z|\bm{x}_n), \qquad \bm{x}_n \sim p(\bm{x}),
\end{align}
and
\begin{align}
    \mathrm{KLD}(p||q) &= \mathbb{E}_p \left[ \ln \frac{p(z)}{q(z)} \right] \\
                      & \cong \sum_{l}(\ln p(z_l) - ln q(z_l)), \qquad z_l \sim p
\end{align}
we write
\begin{align}
    \mathrm{KLD} \left( \sum_{n=1}^{N} p(z|\bm{x}_n), p(z) \right) &\cong \frac{1}{L}\sum_{l}^{L}\left\{\underbrace{\ln\left[\sum_{n=1}^{N} p(z_l|\bm{x}_n)\right]}_{:=LNP}\underbrace{- \ln[\mathcal{N}(0, I)]  }_{:=LNQ}\right\}
\end{align}

where $\mu_n = \mu(\bm{x}_n)$ and $\Sigma_n = \Sigma(\bm{x}_n)$.
Furthermore,

\begin{align}
LNP &= \ln \left[ \sum_{n} \frac{1}{\sqrt{\left( (2\pi)^s |\Sigma_n|\right)}} e^{-\frac{1}{2}(z_l - \mu_n)^T \Sigma_n^{-1}(z_l-\mu_n)}    \right] 
\end{align}

and
\begin{equation}
LNQ = - \left[ -\frac{s}{2} \ln(2 \pi) - \frac{z^Tz}{2} \right] = \frac{1}{2}\left[s \ln(2\pi) + z^T z \right],
\end{equation}
with $s$ equal to the dimensionality of the latent space.

\newpage

\section*{Appendix C}
\subsection*{Robustness to degradation in input imaging data}
\begin{figure}
    \centering
    \includegraphics[width=0.7\textwidth]{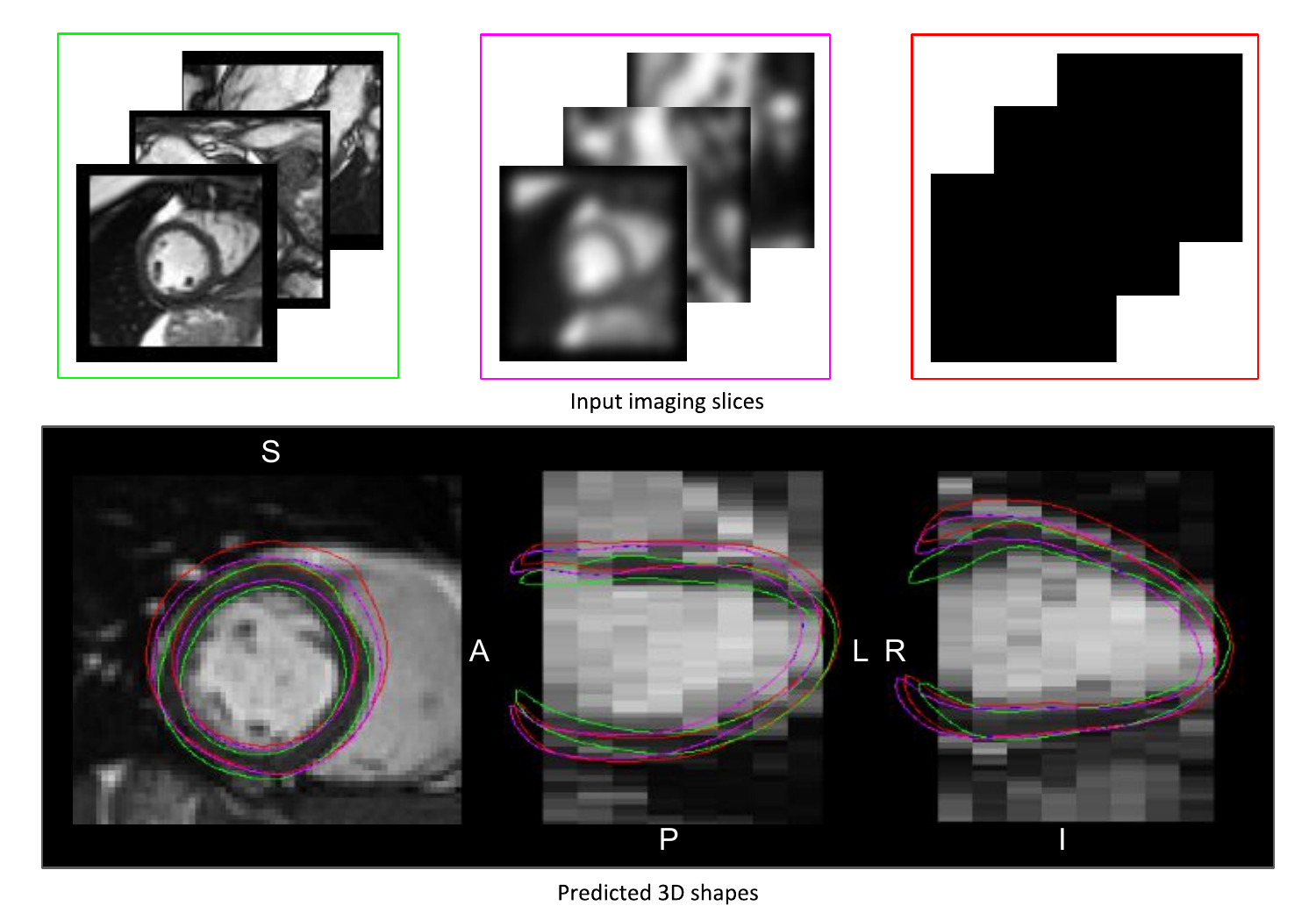}
    \caption{Qualitative assessment. Prediction constructed using degrading input images whilst keeping the input coordinate maps intact. Whilst positioned correctly, the predicted shape no longer exactly corresponds to the original prediction in the presence of image degradation. While the input coordinate maps provide proper localisation in the world space, the input MR imaging slices refine the shape so that it relates to the underlying intensity information. At the extreme - when the input image slices carry no image information (the black slices), the predicted shape still looks like a myocardium of the left ventricle. This is due to the fact that by design the method learns to predict only elements on the manifold of the permissible shapes.}
    \label{fig:robustness}
\end{figure}

\end{document}